%% file: main.tex
\documentclass[10pt]{article} 
\usepackage[accepted]{tmlr}

\input{math_commands.tex}

\usepackage[dvipsnames]{xcolor}
\usepackage[table]{xcolor}  
\usepackage{hyperref}
\usepackage{url}
\usepackage[utf8]{inputenc} 
\usepackage[T1]{fontenc}    
\usepackage{booktabs}       
\usepackage{amsfonts}       
\usepackage{nicefrac}       
\usepackage{microtype}      
\usepackage{enumitem}

\usepackage{graphicx}
\usepackage{wrapfig}
\usepackage{subcaption}
\usepackage{comment}
\usepackage{makecell}
\usepackage[nameinlink]{cleveref}

\Crefname{equation}{Eq.\!}{Eqs.\!}
\Crefname{figure}{Fig.\!}{Figs.\!}
\creflabelformat{equation}{#2#1#3}
\crefrangelabelformat{equation}{#3#1#4-#5#2#6}
\definecolor{paramcolor}{HTML}{41AB5D}

\usepackage{soul}
\usepackage{amsmath}
\usepackage{amssymb}
\usepackage{mathtools}
\usepackage{amsthm}
\usepackage{nicefrac}
\usepackage{algorithm,algpseudocode}
\usepackage{bm}
\usepackage{xcolor}
\usepackage{multirow}

\title{Improved denoising diffusion probabilistic models with efficient non-diagonal covariance modeling}

\author{\name Rui Xia \email rx220@cam.ac.uk \\
\addr University of Cambridge, UK
\AND 
\name Ayan Das$^*$ \email ayan.das@kcl.ac.uk\\
\addr King's College London, UK
\AND 
\name Artem Artemev \email art.art.v@gmail.com\\
\addr MediaTek Research, UK 
\AND 
\name Andi Zhang$^*$ \email andi.zhang@warwick.ac.uk\\
\addr University of Warwick, UK
\AND 
\name Guillaume Hennequin \email gjeh2@cam.ac.uk \\
\addr University of Cambridge, UK
\AND
\name Alberto Bernacchia \email alberto.bernacchia@mtkresearch.com\\
\addr MediaTek Research, UK
}

\def\month{06}  
\def\year{2026} 
\def\openreview{\url{https://openreview.net/forum?id=V6FBm4kfML}} 

\begin{document}

\maketitle

\begin{abstract}
The sampling process of Denoising Diffusion Probabilistic Models (DDPMs) can be accelerated by leveraging second-order information in the form of approximations to the denoising posterior covariance -- allowing samples of acceptable quality to be produced in fewer but larger sampling steps.
  Previous attempts at using such information have used drastic (e.g.\ diagonal) simplifications of the covariance.
  These do not do justice to the peculiar statistical structure of natural images, which exhibit strong non-diagonal correlations between pixels and color channels, and a slow-decaying power-law frequency spectrum. 
  Here, we develop a novel covariance model that captures these features. Our Kronecker-DCT (K-DCT) model uses a Kronecker-factored decomposition of inter-color  covariances and spatial covariances modeled in the frequency domain using the Discrete Cosine Transform (DCT).
  The use of the DCT reduces the computational complexity from quadratic to log-linear, resulting in negligible computational and memory overhead in each denoising step.
  By learning K-DCT-structured amortizations of the denoising posterior covariance using pre-trained score models on CIFAR-10, Celeb-A, ImageNet and LSUN datasets, we show improved performance compared to previous SOTA denoising samplers, both in terms of FID and likelihoods, especially in the regime of few denoising steps.
\end{abstract}

\section{Introduction}\label{sec:intro}

Denoising Diffusion Probabilistic Models (DDPMs; \citealp{ho2020denoising,song2020score,turner2024denoising}) are a family of generative models used ubiquitously for image generation \citep{rombach2022high,esser2024scaling,dalle-3},
where they give state-of-the-art performance both in terms of fidelity (quality of samples) and mode-coverage (sample diversity).
These models sample new images by running a so-called `denoising' Markov chain, starting from pure noise. Given the current image $\bm{x}_t$ at time step $t$, a slightly less noisy image $\bm{x}_{t-\delta}$ is obtained by sampling from a Gaussian approximation to the `denoising posterior' $p(\bm{x}_{t-\delta} | \bm{x}_t)$, under a certain probabilistic model that defines their joint distribution.
Most research efforts so far have focused on approximating the first moment of this posterior, i.e.\ the conditional mean $\mathbb{E}[\bm{x}_{t-\delta}|\bm{x}_t]$, using deep networks trained through various objectives.
The main justification for not paying much attention to the second-order moment (i.e. $\text{Cov}[\bm{x}_{t-\delta} | \bm{x}_t]$) is that, with enough, and small enough, denoising steps, the posterior covariance has a simple (diagonal) form available in closed-form \citep{anderson1982reverse,
song2020score}.
However, taking many small steps in an inherently sequential algorithm is not easily parallelized, such that a trade-off arises between sample quality and sampling time. 

To speed up image generation, one can formulate
diffusion in the (smaller) latent space of a pretrained image autoencoder \citep{rombach2022high}, express the stochastic denoising process as an equivalent deterministic ODE that can be accelerated by appropriate choices of (e.g. higher-order) ODE solvers \citep{song2020score,dpm_solver,elucidate,zheng2023dpm, zhou2024fast, chen2024trajectory}, or outright distill the sampling process into a one-step network \citep{luo2023diff,zhou2024score}. 
Although distilled one-step samplers dominate current practice by offering fast, high-quality generation, they lose the benefit of DDPM's tractable likelihood estimation.
Here, we follow another line of recent research that has shown that standard DDPM sampling can be accelerated by performing fewer but larger steps.
This increased efficiency is achieved by using a more accurate model of the posterior covariance \citep{baoanalytic,bao2022estimating,nichol2021improved,rissanen2025free}. As it turns out, any score network that has been (well) trained to approximate the posterior mean contains all the information needed to estimate the covariance, too.
This relationship has been formalized recently through a generalization of Tweedie's formula \citep{efron2011tweedie} to higher order moments \citep{manorposterior}, revealing an analytical relation between high-order posterior moments and derivatives of the posterior mean (or, alternatively, of the `score' function).
This second-order information contained in pre-trained diffusion models can be distilled into parametric models of the covariance, either by differentiating through the score network exactly \citep{ou2024diffusion} or approximately \citep{manorposterior}, or by reformulating the posterior covariance as the minimum mean squared error (MSE) estimator of a quantity involving the posterior mean (\citealp{meng2021higherorder}; see also \hyperref[sec:background]{Background}).
However, for models that generate color images with $D = d^2$ pixels, the full posterior covariance matrix (or, equivalently, its square root) has a large memory footprint ($3D \times 3D$) implying $\mathcal{O}(D^2)$ sampling complexity, calling for more tractable approximations.
This tradeoff is not unlike that encountered in neural network optimization, where accurately modeling the (second-order) curvature of the loss enables the use of larger learning rates, yet loss Hessians are large objects that can only be estimated in approximate, memory-efficient forms \citep{martens2015optimizing,garcia2023fisher,goldfarb2020practical}. \citet{rissanen2025free} have recently leveraged this connection to improve image restoration.

\begin{figure}[t]
    \centering
    \includegraphics[width=\linewidth]{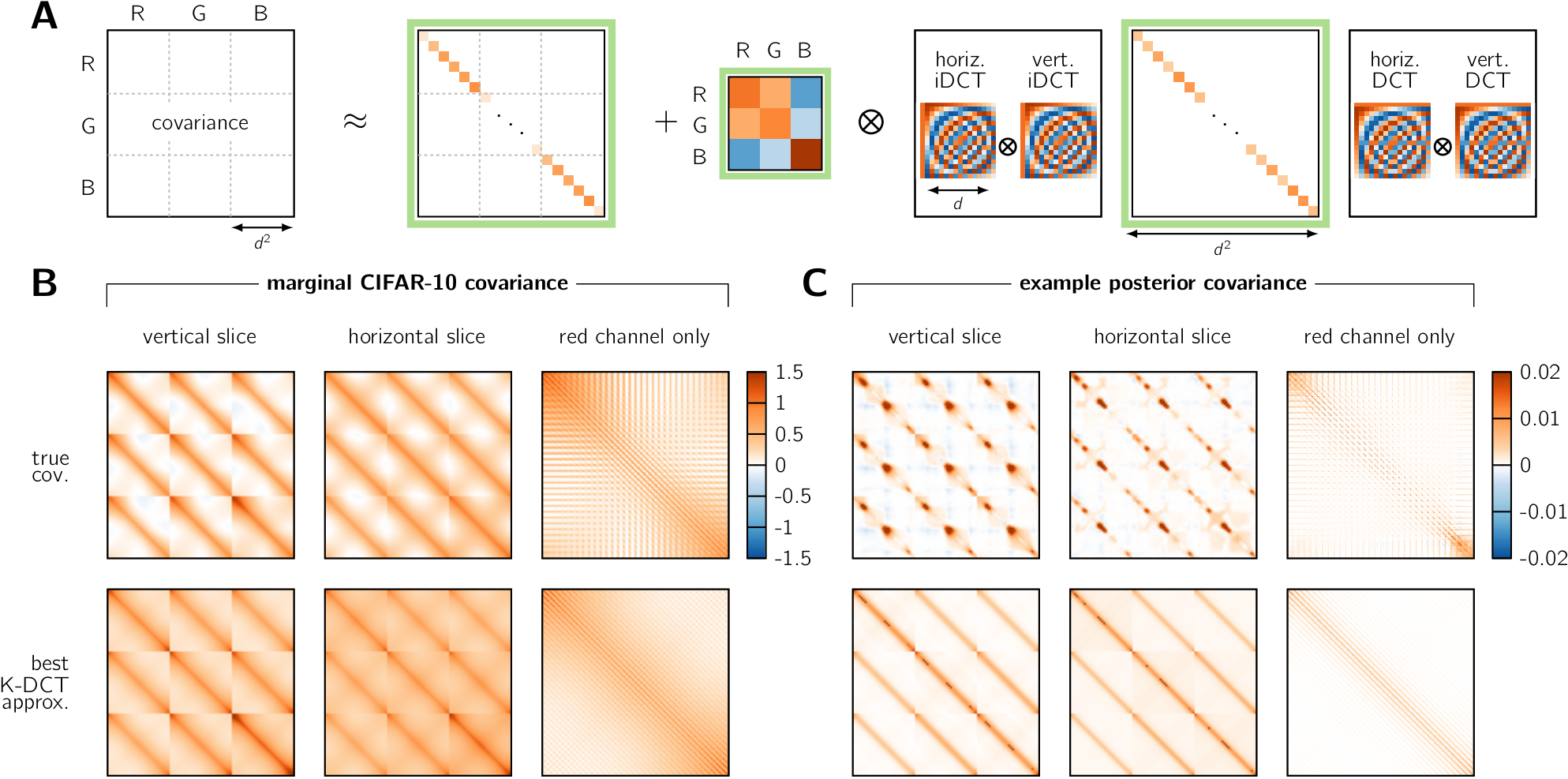}
    \caption{\textbf{Covariance matrices for denoising diffusion probabilistic models are well approximated by a Kronecker-DCT (K-DCT) structure.}
    (\textbf{A})~Illustration of our K-DCT covariance approximation (\Cref{eq:our_param}), with learnable parameters indicated in green. $\otimes$ denotes the Kronecker product.
    (\textbf{B})~Top: covariance of the CIFAR-10 dataset (image width $d=32$), shown across RGB channels ($3 \times 3$ block structure) but restricted to pixels along the vertical (left) and horizontal (center) image midlines, and shown in full ($d \times d$ block structure, right) but for the red channel only. 
    Bottom: same visualizations of the nearest (in minimum squared error sense) approximation of the CIFAR-10 covariance that conforms to the structure shown in (A).
    See also \Cref{fig:kronecker-imagenet,fig:kronecker-celeba} for further analyses of how \Cref{eq:our_param} accurately describes the marginal covariance structure of larger datasets (ImageNet and CelebA), see \Cref{fig:suppl:audio} for other data modality, speech data.
    (\textbf{C})~Same as (B), for an example posterior covariance matrix obtained from the Jacobian of a score network (see \Cref{eq:tweedie_2}) pre-trained on CIFAR-10 \citep{bao2022estimating}, evaluated at a partially denoised sample ($600$ denoising steps, i.e.\ roughly mid-way through denoising).See also \Cref{fig:kronecker-cifar} for a further dissection of how \Cref{eq:our_param} performs at various stages of denoising on CIFAR-10.
    }
    \label{fig:illustration}
\end{figure}

All recent attempts at modeling denoising posterior covariances have assumed a diagonal or low-rank structure, which we argue is very restrictive.  
Here, we develop a new covariance model for image DDPMs (\Cref{fig:illustration}A) which accurately and efficiently captures the strong yet non-diagonal spatio-chromatic correlations between both neighbouring pixels and color channels present in natural images (\Cref{fig:illustration}B; \citealp{burton1987color, cui2020color, fairman2004principal}).
These chromatic and spatial correlations are approximately separable \citep{provenzi2016second}, and therefore Kronecker-factorizable, and the spatial component can be compactly represented in the frequency domain of the Discrete Cosine Transform (DCT) owing to approximate translation invariance \citep{hyvarinen2009natural}. The resulting `K-DCT' model is described in detail in \Cref{sec:kdct}; \Cref{fig:illustration}B (bottom) shows that it provides a good fit to the marginal (i.e.\ prior) CIFAR-10 covariance, and this paper explores its use for approximating the posterior covariances that arise in image denoising -- an example of which is shown in \Cref{fig:illustration}C (top) along with its best K-DCT approximation (bottom).
Starting from pre-trained score models, we learn K-DCT-structured amortizations of the (input-dependent) posterior covariance $\text{Cov}(\bm{x}_{t-\delta} | \bm{x}_t) \approx \text{K-DCT}(\bm{x}_t;\theta)$.
On CIFAR-10, Celeb-A,  ImageNet and LSUN, we show that in the regime of few sampling steps, this leads to both better image generation (lower FID; \citealp{heusel2017gans}) and better statistical models (lower negative log-likelihood) compared to previous diagonal approximations.

\section{Background}
\label{sec:background}

In this section, we begin by providing important background on the general Gaussian denoising problem, highlighting two ways of obtaining the mean and covariance of the denoising posterior distribution.
We then discuss how these two denoising strategies can be applied to the sampling process in DDPMs, which we also summarize.

\subsection{The Gaussian denoising problem}
\label{sec:denoising}

Consider a random vector $\bm{x} \in \mathbb{R}^n$ drawn from some distribution $q(\bm{x})$.
Given a noisy observation $\tilde{\bm{x}} \sim q(\tilde{\bm{x}} | \bm{x}) = \mathcal{N}(\tilde{\bm{x}}; \bm{x}, \sigma^2 I)$, what can be said about $\bm{x}$?
Whilst the posterior distribution $q(\bm{x} | \tilde{\bm{x}})$ is generally intractable (e.g.\ the prior $q(\bm{x})$ may not be Gaussian), there are at least two ways of obtaining its moments.

\paragraph{Posterior moments via Tweedie's 1$^\text{st}$- and 2$^\text{nd}$-order formulae}
Posterior moments can be derived from the score function, 
$\nabla_{\tilde{\bm{x}}} \log \tilde{q}(\tilde{\bm{x}})$, where $\tilde{q}(\tilde{\bm{x}}) = \int d\bm{x} \, q(\tilde{\bm{x}}|\bm{x}) \, q(\bm{x})$ is the marginal distribution of noisy observations.
Tweedie's formula \citep{efron2011tweedie,robbins1992empirical} classically relates the posterior mean to the score function:
\begin{equation}
    \label{eq:tweedie_1}
    \mu^\star(\tilde{\bm{x}}) \triangleq \mathbb{E}[\bm{x}|\tilde{\bm{x}}] = \tilde{\bm{x}} + \sigma^2 \nabla_{\tilde{\bm{x}}} \log \tilde{q}(\tilde{\bm{x}}).
\end{equation}
A similar relationship exists between the posterior covariance and the \emph{second} derivative of $\log \tilde{q}(\cdot)$ (i.e.\ the Jacobian of the score; \citealp{manorposterior,meng2021higherorder}):
\begin{equation}
  \Sigma^\star(\tilde{\bm{x}}) \triangleq \text{Cov}[\bm{x}|\tilde{\bm{x}}] = \sigma^2 \nabla_{\tilde{\bm{x}}} \mu^\star(\tilde{\bm{x}}) \nonumber = \sigma^2 \left(I + \sigma^2 \nabla^2_{\tilde{\bm{x}}} \log \tilde{q}(\tilde{\bm{x}})\right).
  \label{eq:tweedie_2}
\end{equation}

\paragraph{Posterior moments as least-squares estimators}
When one has access to $(\bm{x}, \tilde{\bm{x}})$ pairs (e.g.\ via simulation: $\bm{x} \sim q(\bm{x})$, $\tilde{\bm{x}} \sim q(\tilde{\bm{x}} | \bm{x})$), one can estimate posterior moments from data by minimizing a squared error loss.
Indeed, for any deterministic function $g(\cdot)$, the conditional expectation of $g(\bm{x})$ under the posterior $q(\bm{x}|\tilde{\bm{x}})$ is the solution to a mean-squared error minimization problem:
\begin{equation}
    \mathbb{E}_{q(\bm{x}|\tilde{\bm{x}})}[g(\bm{x})] =
    \text{argmin}_{y(\cdot)} \ \mathbb{E}_{q(\tilde{\bm{x}}|\bm{x})q(\bm{x})} \left[ \| y(\tilde{\bm{x}}) -  g(\bm{x}) \|^2 \right].
    \label{eq:mmse_estimation_obj}
\end{equation}
Thus, one can estimate the posterior mean function (i.e. $g(\bm{x}) = \bm{x}$), in parametric form $\mu_\theta(\tilde{\bm{x}})$, by minimizing
\begin{equation}
   \theta^\star = \text{argmin}_\theta \ \mathbb{E}_{q(\tilde{\bm{x}}|\bm{x})q(\bm{x})} \left[ \| \mu_\theta(\tilde{\bm{x}}) -  \bm{x} \|^2 \right].
    \label{eq:mmse_mean}
\end{equation}
with the expectation typically estimated via Monte-Carlo sampling
of $(\bm{x},\tilde{\bm{x}})$ pairs. 
Similarly, a parametric posterior covariance model $\Sigma_\theta(\tilde{\bm{x}})$ can be learned by minimizing
\begin{equation} 
   \theta^\star = \text{argmin}_\theta \ \mathbb{E}_{q(\tilde{\bm{x}}|\bm{x})q(\bm{x})} \left[ \left\| \Sigma_\theta(\tilde{\bm{x}}) -  (\bm{x} - \boldsymbol\mu^\star(\tilde{\bm{x}}))
   (\bm{x} - \boldsymbol\mu^\star(\tilde{\bm{x}}))^\top \right\|^2 \right]
    \label{eq:mmse_cov} \\
\end{equation}
with $\boldsymbol\mu^\star\!$ either derived from the score function through \Cref{eq:tweedie_1}, or parametrically estimated using \Cref{eq:mmse_mean}.

\subsection{Denoising diffusion probabilistic models}
\label{sec:ddpm}

DDPMs \citep{ho2020denoising} are probabilistic models that allow sampling from an arbitrary distribution $q(\bm{x}_0)$.
Just as in the Gaussian denoising problem discussed above, DDPMs define a whole collection of noisy observations $\{\bm{x}_1, \ldots, \bm{x}_T\}$, obtained by sequentially down-scaling, and adding Gaussian noise to, each data sample $\bm{x}_0$:
$
\bm{x}_t = \sqrt{\alpha_t} \bm{x}_{t-1} + \sqrt{\beta_t} \boldsymbol\epsilon_t \quad \text{with} \quad
\boldsymbol\epsilon_t \sim \mathcal{N}(0,I),
$
where $\beta_t$ is a time-dependent diffusion coefficient also known as the `noising schedule'.
Typically, $\alpha_t = 1-\beta_t$, a choice that preserves the total variance of $\bm{x}_t$ at each step $t$.
Having introduced this forward `noising' Markov chain, the data distribution can be expressed as
$
q(\bm{x}_0) = \int d\{\bm{x}_1, \ldots, \bm{x}_T\} q(\bm{x}_T) \prod_{t=1}^T q(\bm{x}_{t-1} | \bm{x}_t).
$
Therefore, sampling from $q(\bm{x}_0)$ can be achieved by sampling from $q(\bm{x}_T)$ and then running a sequence of small denoising steps whereby each $\bm{x}_{t-1}$ is obtained from $\bm{x}_t$ by sampling the relevant denoising posterior $q(\bm{x}_{t-1} | \bm{x}_t)$.
Critically, for a sufficiently long noising process, $\bm{x}_T$ is approximately normally distributed and is therefore trivial to sample.

In general, each posterior $q(\bm{x}_{t-1}|\bm{x}_t)$ is intractable, and is normally approximated by a Gaussian:
\begin{equation}
   \label{eq:approx_posterior} 
   q(\bm{x}_{t-1} | \bm{x}_t) \approx \mathcal{N}\left(\bm{x}_{t-1}; 
        \bm\mu_{t-1}(\bm{x}_t), \Sigma_{t-1}(\bm{x}_t)\right).
\end{equation}
Note that it is possible to generate a sample $\bm{x}_0$ using a number of denoising steps smaller than $T$, by merging any number of consecutive denoising steps into a single one (`skip-step DDPM').
This is done by leveraging the fact that, under the noising model, any $\bm{x}_t$ is a linear-Gaussian transformation not only of $\bm{x}_{t-1}$ but of any previous $\bm{x}_{s<t}$.
This leads to simple affine relationships between $\{\bm\mu_{t-1}(\bm{x}_t), \Sigma_{t-1}(\bm{x}_t)\}$ and the more general $\{\bm\mu_s(\bm{x}_t), \Sigma_s(\bm{x}_t)\}$ (\Cref{sec:skipstep}).
It is precisely when skipping steps that the posterior covariance becomes less diagonally dominant, such that it becomes important to accurately model its structure -- the focus of this paper.
We now discuss how estimates of $\boldsymbol\mu_{t-1}(\bm{x}_t)$ and $\Sigma_{t-1}(\bm{x}_t)$ can be obtained.

\paragraph{Posterior mean}
The posterior mean function $\boldsymbol\mu_{t-1}(x_t)$ is typically obtained indirectly by estimating the \emph{effective noise} term $\boldsymbol\epsilon_t \triangleq \frac{\bm{x}_t - \sqrt{\bar\alpha_t} \bm{x}_0}{\sqrt{\bar\beta_t}}$ that transformed $\bm{x}_0$ into $\bm{x}_t$, using standard notation $\bar\alpha_t \triangleq \prod_{s=0}^t \alpha_s$ and $\bar\beta_t \triangleq 1 - \bar\alpha_t$.
Indeed, simple affine transformations exist between the conditional expectation $\mathbb{E}[\boldsymbol\epsilon_t | \bm{x}_t]$ and $\mathbb{E}[\bm{x}_0 | \bm{x}_t]$, and further towards $\boldsymbol\mu_{t-1}(\bm{x}_t) 
 \equiv \mathbb{E}[\bm{x}_{t-1} | \bm{x}_t]$ as follow:
\begin{equation}
   \label{eq:affine}
   \mathbb{E}[\bm{x}_0 | \bm{x}_t] = \frac{
    \bm{x}_t -  \sqrt{\bar\beta_t} \mathbb{E}[\boldsymbol\epsilon_t | \bm{x}_t]}{\sqrt{\bar\alpha_t}}, \,\quad \boldsymbol\mu_{t-1}(\bm{x}_t) = \frac{
    \sqrt{\bar\alpha_{t-1}} \, \beta_t \mathbb{E}[\bm{x}_0 | \bm{x}_t]
    + \sqrt{\alpha_t} \bar\beta_{t-1} \bm{x}_t
    }{\bar\beta_t}.
\end{equation}
In practice, a neural network $\boldsymbol\epsilon_\theta(\bm{x}_t,t)$ is trained to approximate $\mathbb{E}[\boldsymbol\epsilon_t | \bm{x}_t]$, and used to evaluate
\Cref{eq:affine} (see also 
\Cref{sec:detail_training} for more practical details, including the image-specific use of clipping).

\paragraph{Posterior covariance}
The posterior covariance function, $\Sigma_{t-1}(\bm{x}_t)$, is often approximated by one of two time-dependent, but $\bm{x}_t$-independent, heuristics: $\beta_t I$ (`large'), or $\tilde\beta_t I$ (`small') with $\tilde\beta_t \triangleq \frac{1-\bar\alpha_{t-1}}{1-\bar\alpha_t} \beta_t$.
These become equal, and exact, in the limit of many small (de-)noising steps, i.e.\ a limit where $\bm{x}_0$ contains much less information about $\bm{x}_{t-1}$ than does $\bm{x}_t$. 
For realistically small horizons $T$, however, the posterior covariance may be far from being a scalar, or even a diagonal matrix (e.g.\ \Cref{fig:illustration}C).
Several works have sought to learn better models of the posterior covariance in parametric form.   
Similarly to the posterior mean function, the posterior covariance function $\Sigma_{t-1}(\bm{x}_t)$ is mathematically related to the covariance of the noise, $\text{Cov}[\bm{\epsilon}_t|\bm{x}_t]$, as follows: 
\begin{equation}
    \label{eq:affinecov}
    \text{Cov}[\bm{x}_0|\bm{x}_t] = \frac{\bar\beta_t}{\bar\alpha_t}\text{Cov}[\bm\epsilon_t|\bm{x}_t], \, \,
    \Sigma_{t-1}(\bm{x}_t) = \tilde{\beta}_t I + \frac{\beta_t^2\bar\alpha_{t-1}}{(1-\bar\alpha_t)^2}\text{Cov}[\bm{x}_0|\bm{x}_t].
\end{equation}

Hence, to perform the denoising sampling step of \Cref{eq:approx_posterior} given a pretrained model that already approximates the posterior mean of the effective noise term, it is sufficient to learn a parametric model $\mathcal{E}_\phi(\bm{x}_t,t)$ of $\text{Cov}(\boldsymbol{\epsilon}_t| \bm{x}_t)$.
This model needs to have a manageable memory footprint, and its matrix square root (required for sampling) must afford computationally tractable matrix-vector products.
The K-DCT covariance model we propose here is equally applicable to the two ways of obtaining posterior covariances described in \Cref{sec:denoising}: either via derivatives of the score function (Tweedie's 2$^\text{nd}$-order formula), or via direct least-squares estimation from data.
In the following, we describe their specific application to the DDPM denoising posterior.

\subsection{Learning posterior covariance approximations for DDPMs}

\paragraph{Score derivative-based approach}
Leveraging the connection between the denoising posterior covariance and the Jacobian of the score \citep{manorposterior}, \citet{ou2024diffusion} learned a parametric diagonal covariance model $\mathcal{E}_\phi(\bm{x}_t,t) = \text{diag}(\boldsymbol\varepsilon_\phi(\bm{x}_t, t))$ by minimizing the following `optimal covariance matching' (OCM) objective:
\begin{equation}
    \label{eq:ocm_objective}
    \mathcal{L}_{\text{OCM}}(\phi) = \mathbb{E}_{q(\bm{x}_0)q(\bm{x}_t|\bm{x}_0)} \left\|
    \boldsymbol\varepsilon_\phi(\bm{x}_t,t) -
    \text{diag}\left(I - \sqrt{\bar\beta_t}\nabla_{\bm{x}_t}\boldsymbol\epsilon_\theta(\bm{x}_t,t) \right) \right\|^2_2,
\end{equation}
where $\boldsymbol\epsilon_\theta(\cdot, \cdot)$ is a pretrained first-order model approximating $\boldsymbol\epsilon_t$ (see \Cref{sec:ddpm}), and $\text{diag}(M)$ extracts the diagonal of matrix $M$.
Here, we have adapted their objective of \Cref{eq:ocm_objective}, which targets the gradient of the score, to an equivalent formulation which targets $\text{Cov}(\boldsymbol{\epsilon}_t | \bm{x}_t)$ instead (\Cref{eq:affinecov,eq:tweedie_2}).
For non-diagonal approximations (such as ours; see below), one cannot afford materializing the residual in \Cref{eq:ocm_objective} in order to compute its squared norm. To circumvent this, \citeauthor{ou2024diffusion} used an unbiased stochastic estimator of the corresponding gradient, obtained by automatically differentiating through the following surrogate objective
\begin{equation}
\tilde{\mathcal{L}}_\text{OCM}(\phi)
= 
\mathbb{E}_{q(\bm{x}_0)q(\bm{x}_t|\bm{x}_0)}
\mathbb{E}_{\bm{v} \sim p(\bm{v})} \left\|
  \boldsymbol\varepsilon_\phi(\bm{x}_t, t) -
  \bm{v} \odot \left( \bm{v} - \sqrt{\bar\beta_t}\nabla_{\bm{x}_t}\boldsymbol\epsilon_\theta(\bm{x}_t,t) \bm{v} \right)
\right\|^2_2,
\end{equation}
where $\odot$ denotes the Hadamard (element-wise) product.
The inner expectation can be stochastically estimated via Monte-Carlo sampling of $\bm{v}$ from an isotropic Rademacher distribution.
Note that the $\nabla_{\bm{x}_t} \boldsymbol\epsilon_\theta(\bm{x}_t, t) \bm{v}$ term is a Jacobian-vector product~(JVP) that can be calculated efficiently using forward-mode auto-differentiation~(AD).
As it does not depend on $\phi$, this JVP needs not be further differentiated (i.e.\ no need for nested AD).
Here, we will adapt this approach to deal with more general, non-diagonal covariance models (\Cref{sec:covofnoiseproof,sec:eff_training_sampling}).

\paragraph{MMSE approach}
\citet{meng2021higherorder} leveraged the least-squares estimator interpretation of the denoising posterior moments (\Cref{sec:denoising}) to learn amortizations of higher-order derivatives of any data (log) distribution.
In turn, they showed that a good second-order score approximation leads to better denoising uncertainty quantification.
More recently, \citet{bao2022estimating} followed a similar approach to fit a parametric model of $\text{Cov}(\boldsymbol{\epsilon}_t| \bm{x}_t)$ in the form $\mathcal{E}_\phi(\bm{x}_t,t) = \text{diag}(\boldsymbol\varepsilon_\phi(\bm{x}_t, t))$,
 using a MMSE objective.
Given independent $(\bm{x}_0, \boldsymbol\epsilon_t)$ pairs and the associated $\bm{x}_t = \sqrt{\bar\alpha_t} \bm{x}_0 + \sqrt{\bar\beta_t} \boldsymbol\epsilon_t$, their `noise prediction residual' (NPR) objective reads
\begin{equation}
    \label{eq:npr_objective}
    \mathcal{L}_{\text{NPR}}(\phi)  = \mathbb{E}_{\bm{x}_0 \sim q(\bm{x}_0); \boldsymbol\epsilon_t \sim \mathcal{N}(0, I)} \left\|
    \boldsymbol\varepsilon_\phi(\bm{x}_t,t)- \left(\boldsymbol\epsilon_t - \boldsymbol\epsilon_\theta(\bm{x}_t,t)\right)^2 \right\|^2_2.
\end{equation}
This objective again relies on a pretrained first-order noise predictor $\boldsymbol\epsilon_\theta(\bm{x}_t,\!t)$.
Similar to the OCM objective, we will adapt the NPR objective to deal with more general, non-diagonal covariance models.

\section{Covariance Parameterizations}

Tractable evaluation and differentiation of both the OCM (\Cref{eq:ocm_objective}) and NPR (\Cref{eq:npr_objective}) objectives  places constraints on the form of covariance approximation ($\mathcal{E}_\phi(\cdot)$) that may be used.
One highly flexible, but also highly intractable, choice would be to parameterize the Cholesky factor of the entire $3D \times 3D$ covariance matrix, where $D=d^2$ is the number of image pixels -- this leads to a prohibitive $\mathcal{O}(D^2)$ memory and compute complexity.
In this work, we introduce a model $\mathcal{E}_{\phi}$ that provides not only a better inductive bias for image generation than previous proposals (briefly reviewed below), whilst affording efficient training and sampling.

\subsection{Existing parameterizations}

As previously mentioned, a popular covariance approximation is the diagonal parameterization: e.g.\ \citet{iddpm} and \citet{ou2024diffusion} parameterize $\boldsymbol\varepsilon_\phi(\bm{x}_t, t) \in \mathbb{R}^{3D}$ such that $\mathcal{E}^\text{diag}_\phi(\bm{x}_t, t) = \text{diag}(\boldsymbol\varepsilon_\phi(\bm{x}_t, t)) \in \mathbb{R}^{3D\times 3D}$.
This has $\mathcal{O}(D)$ (linear) complexity in both training and sampling, but fails to take into account pairwise correlations between pixels.
To capture the dominant patterns of pairwise correlations under the denoising posterior,  \citet{meng2021higherorder} added a low-rank component to the diagonal, resulting in:
\begin{equation}
\mathcal{E}_\phi(\bm{x}_t, t) = \text{diag}(\boldsymbol\varepsilon_\phi(\bm{x}_t, t)) + R_\phi(\bm{x}_t, t)R_\phi(\bm{x}_t, t)^\top 
\end{equation}
where $R_\phi \in \mathbb{R}^{3D\times r}$ with $r \ll 3D$. However, we find that the eigenvalue spectra of image denoising posterior covariances tend to decay slowly
(see also \citealp{van1996modelling}), such that $r$ might need to be fairly large to capture useful structure.
This is corroborated by \citeauthor{meng2021higherorder}'s qualitative results on the MNIST dataset, where they used $r=50$ (i.e.\ 6\% of $D$) to obtain a posterior covariance approximation that contained the patterns of denoising uncertainty between digits that one would intuitively expect.
In the CIFAR-10 example of \Cref{fig:illustration}C, capturing 90\% of the variance in the denoising covariance matrix requires setting $r=635 \approx 20\% \text{ of }3D$.
For larger images, such as those in the datasets we consider here (CelebA and ImageNet), better forms of approximation are needed that can capture the full rank of the posterior covariance without introducing an additional compute/memory tradeoff.

\subsection{Proposed K-DCT parameterization}
\label{sec:kdct}

Motivated by key statistical properties of natural images (recall \hyperref[sec:intro]{Introduction} and \Cref{fig:illustration}) and by the spectral theory of discrete cosine transforms, we propose the following covariance model for image DDPMs, with individual components explained in detail below:
\begin{equation}
\mathcal{E}_\phi(\bm{x}_t, t) =
\underbrace{\text{diag}(\boldsymbol\varepsilon_\phi(\bm{x}_t, t))}_\text{diagonal baseline}
+ \underbrace{C_\phi(\bm{x}_t, t) C_\phi(\bm{x}_t, t)^\top}_\text{inter-channel} 
\otimes \underbrace{(\overbrace{F^\top}^\text{horiz.} \!\otimes \! \overbrace{F^\top}^\text{vert.}) \text{diag}(\bm\lambda_\phi(\bm{x}_t, t)) \left(F\!\otimes\! F\right)}_\text{inter-pixel}.
\label{eq:our_param}
\end{equation}
%
%
\setlength{\leftmargini}{0.4cm}
\begin{itemize}
    \item \textbf{Diagonal baseline} -- This term absorbs any diagonal contribution that the second (Kronecker-DCT) term might not capture, thereby ensuring that the model is at least as expressive as previous diagonal models we compare to.
    \item \textbf{Outer Kronecker product} -- The second term models the approximately \emph{separable} spatio-chromatic correlation structure of natural images, whereby e.g.\ the red and blue content of two pixels are correlated in the same way irrespective of where these two pixels are located \citep{provenzi2016second}.
    Separability is achieved through a Kronecker product ($\otimes$) of inter-channel (color) correlations and inter-pixel (spatial) correlations, with each component modeled as follows:
    \begin{itemize}
        \item \textbf{Inter-channel} ($3 \times 3$) -- The correlation between RGB channels is captured in full by the $C_\phi^{\vphantom{\top}} C_\phi^\top \in \mathbb{R}^{3 \times 3}$ term.
        \item \textbf{Inter-pixel} ($d^2 \times d^2$) -- For the spatial component, we reason that natural images -- seen as continuous functions of the infinite plane -- have an approximately translation invariant distribution. Thus, their covariance operator has the Fourier modes as eigenfunctions \citep{hyvarinen2009natural}. For discretized (finite-size) images, the first practical parameterization that comes to mind is a diagonal matrix in the orthonormal basis given by the Discrete Fourier Transform (DFT) matrix. However, the DFT assumes cyclic boundary conditions which natural bounded images do not have -- in other words, their covariance is not circulant (as a DFT-based model would assume) but rather Toeplitz \citep{rissanen2025free}. In fact, empirically we find that the CIFAR-10 marginal covariance, in both the horizontal and vertical image directions, is the superposition of a Toeplitz component and a Hankel (``90-deg rotated Toeplitz'') component (\Cref{fig:illustration}B). This suggests using the discrete cosine transform (DCT) instead: matrices diagonalized by the DCT have indeed been shown theoretically to possess precisely this Toeplitz + Hankel structure \citep{sanchez2002diagonalizing}.\begin{itemize}
            \item \textbf{Eigenbasis} -- We therefore consider a spatial covariance diagonalized by the 2-dimensional DCT operator $F \otimes F$, which applies the standard 1-dimensional DCT operator $F \in \mathbb{R}^{d \times d}$  \citep{strang1999discrete} to both the vertical and horizontal image dimensions (note that $F^{-1} = F^\top$).
            \item \textbf{Eigenvalues} -- The $D$ eigenvalues are parameterized in positive real form as $\bm{\lambda}_\phi(\bm{x}_t, t)$.
        \end{itemize} 
        Visual intuition for the expressiveness of this spatial covariance parameterization is given in \Cref{fig:expressiveness}.
    \end{itemize}
\end{itemize}

This model has a small, $\mathcal{O}(D)$ memory footprint (i.e.\ the size of a single image). In the next two subsections, we show that it is also amenable to efficient, $\mathcal{O}(D \log d)$ training and sampling.

\paragraph{Efficient training}
%
The first step in learning non-diagonal covariance models using the OCM or NPR objectives is to extend their diagonal covariance formulations (\Cref{eq:ocm_objective,eq:npr_objective}) to the more general, non-diagonal case.
The corresponding expressions are provided in  \Cref{eq:ocm_full,eq:npr_full} (\Cref{sec:eff_training_sampling}).
Our parametrization in \Cref{eq:our_param} enables efficient evaluation and differentiation of these objectives, much cheaper than the naive approach ($\mathcal{O}(D \log d)$ instead of $\mathcal{O}(D^2)$ compute complexity).
Indeed, the two most expensive operations are matrix-vector products with $\mathcal{E}_\phi$, and the squared Frobenius norm $\| \mathcal{E}_\phi \|_\text{F}^2$ -- both of which can be computed efficiently due to the tensor product structure of  \Cref{eq:our_param}. 
In particular, for an image $V \in \mathbb{R}^{3 \times d \times d}$, the corresponding matrix-vector product can be computed as follows:
\begin{equation}
    \label{eq:mvp}
    (\mathcal{E}_\phi(\bm{x}_t,t)\text{vec}(V))^{cij} =
\textcolor{paramcolor}{\boldsymbol\varepsilon}^{cij} V^{cij}
+    
(F^\top)^j_{\,\, n}(F^\top)^i_{\,\, m}\bigg(\textcolor{paramcolor}{\bm{\lambda}}^{mn}\overbrace{F^n_{\,\, q}F^m_{\,\,\, p} \underbrace{V^{epq} \left(\textcolor{paramcolor}{C C^\top}\right)^c_{\,\, e}}_{\mathcal{O}(3^2)}}^{\mathcal{O}(D\log d)}\bigg)
\end{equation}
where $\textcolor{paramcolor}{\boldsymbol\varepsilon} \equiv \boldsymbol\varepsilon_\phi(\bm{x}_t, t)$,
$\textcolor{paramcolor}{\bm{\lambda}} \equiv \bm{\lambda}_\phi(\bm{x}_t, t)$
and $\textcolor{paramcolor}{C} \equiv C_\phi(\bm{x}_t, t)$.
In \Cref{eq:mvp}, $\text{vec}(\cdot)$ is the tensor vectorization operation, and we have used standard \href{https://en.wikipedia.org/wiki/Einstein_notation}{Einstein notation} whereby repeated indices occurring in opposite super-/sub-scripts are summed over.
In the context of our training losses, $V$ may be either $\boldsymbol\epsilon_t$, $\boldsymbol\epsilon_\theta(\bm{x}_t,t)$, or Rademacher samples used in stochastic trace estimators.
The above equation allows us to never explicitly construct large $3D \times 3D$ matrices, but instead only perform element-wise multiplication and linear transformations of dimension $d$. Note that in theory, products with the DCT matrix such as $F^m_{\ p} V^{p}$ can be computed in $\mathcal{O}(d \log d)$ complexity, in practice we find that GPU-accelerated matrix multiplications with $F$ are faster.
Pseudocode for efficiently evaluating the training loss is given in \Cref{alg:npr_training} (NPR) and \Cref{alg:ocm_training} (OCM).

\paragraph{Efficient sampling}
While reducing training complexity is desirable, using the covariance model for image generation also requires efficient sampling. Sampling from the denoising posterior (\Cref{eq:approx_posterior}) or the skip-step posterior (\Cref{eq:kstep-mean,eq:kstep-cov}) is traditionally done by multiplying a random normal vector by the matrix square-root of $\mathcal{E}_{\phi}(\bm{x}_t,t)$.
While the latter is difficult to obtain for our proposed parameterization due to the sum in \Cref{eq:our_param}, we can instead sample and add two independent samples from the two corresponding multivariate normal distributions.
For the first (diagonal) term, this is straightforward.
The second term admits a simple matrix square root, 
$\textcolor{paramcolor}{C} \otimes \left((F^\top \otimes F^\top) \text{diag}\left(\textcolor{paramcolor}{\bm{\lambda}}^{1/2} \right)\right)$, such that sampling from the corresponding Gaussian can be written analogously to \Cref{eq:mvp} as:
\begin{equation}
 (F^\top)^j_{\,\, n}(F^\top)^i_{\,\, m}\bigg(\left(\textcolor{paramcolor}{\bm{\lambda}}^\frac{1}{2}\right)^{mn}\xi^{emn}\textcolor{paramcolor}{C}^c_{\,\, e}\bigg)
\end{equation}
where $\xi \!\in\! \mathbb{R}^{3D}$ is a standard random normal vector. 
Pseudocode for sampling is given in \Cref{alg:our_eff_sampling}; note that the covariance $\Sigma_{t-1}(\bm{x}_t)$ from which we must sample is not exactly the same as the covariance of the noise ($\mathcal{E}_\phi$) for which pseudocode is given, but it has the same structure (c.f.\ \Cref{eq:affinecov}). 

Empirical comparison of training and sampling time cost between diagonal covariance and our K-DCT model is provided in \Cref{tab:cost}, which shows little computation overhead for our parameterization.

\begin{algorithm}[h]
    \caption{Sampling from $\mathcal{E}_{\phi}(\bm{x}_t,t)$}
    \label{alg:our_eff_sampling}
    \begin{algorithmic}[1]
       \Require Covariance model components $\{ \boldsymbol\varepsilon_{\phi}, C_{\phi}, \bm{d}_{\phi} \}$ with trained parameter set $\phi$, partially denoised $\bm{x}_t$ at a given $t$, two independent Gaussian samples $\boldsymbol\xi_1, \boldsymbol\xi_2\sim\!\mathcal{N}(0,I)$.
       
       \Ensure $\bm{g}$ is a sample from $\mathcal{N}\left(0, \mathcal{E}_\phi(\bm{x}_t,t)\right)$
       \State Compute model outputs $\boldsymbol\varepsilon \!\gets\! \boldsymbol\varepsilon_{\phi}(\bm{x}_t, t)$,
       $C\!\gets C_{\phi}(\bm{x}_t, t)$ and $\bm{d} \gets \bm{d}_{\phi}(\bm{x}_t, t)$
       \State Compute $\tilde{\boldsymbol\xi}_1 \gets \boldsymbol\varepsilon^\frac{1}{2}\odot\boldsymbol\xi_1$ \text{\ \ \# diagonal part; $\odot$ denotes the element-wise product}
       \State Compute $\tilde{\boldsymbol\xi}_2 \gets \text{2D-iDCT}(\text{einsum}(\bm{d}^\frac{1}{2}, C, \boldsymbol\xi_2, \text{\texttt{`ij,ck,kij->cij'}}  )$  \text{\ \ \# non-diagonal part} 
        \State Compute $\bm{g} \gets \tilde{\boldsymbol\xi}_1 + \tilde{\boldsymbol\xi}_2$
    \end{algorithmic}
\end{algorithm}

\section{Experiments \& Results}
\label{sec:exp}
\begin{figure}
    \centering
    \includegraphics[width=\linewidth]{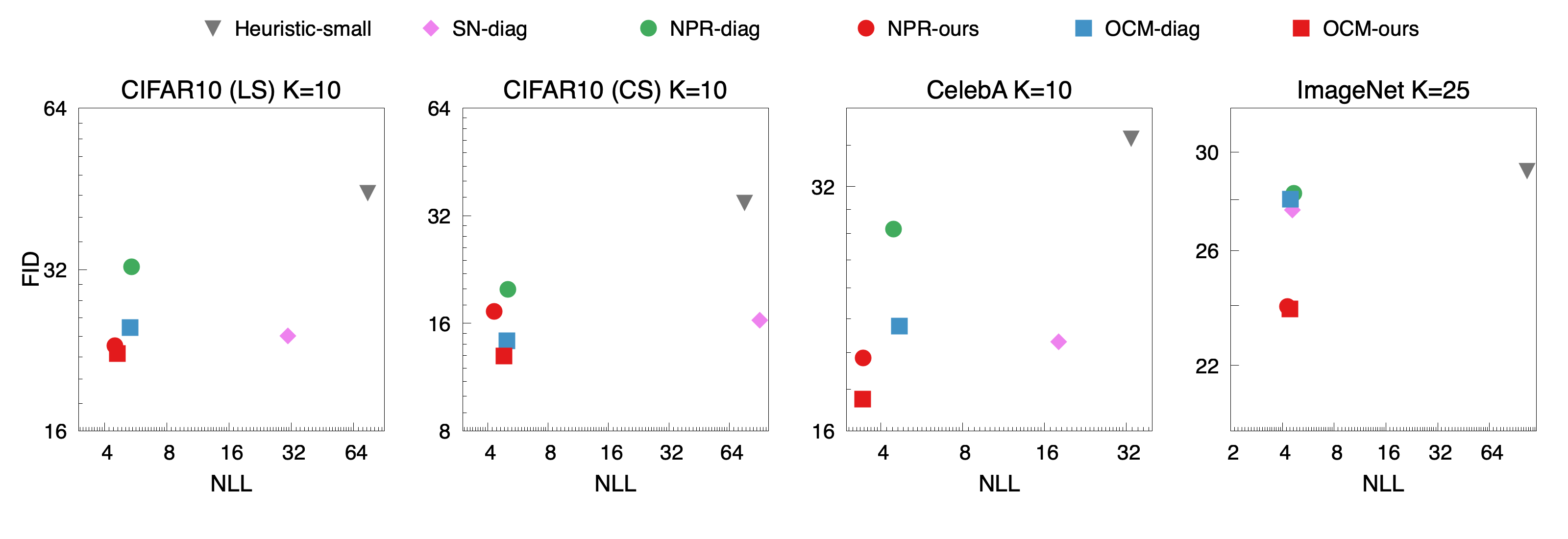}
    \caption{FID (log-scale) vs.\ NLL (log-scale) for different training objectives (OCM, NPR and SN) and different covariance models (heuristic, diagonal, ours) in the skip-step regime of few sampling steps ($K$). Our model consistently achieves both lower FID and NLL.}
    \label{fig:both-better}
\end{figure}

In this section, we run experiments to validate our hypothesis that the K-DCT covariance model (\Cref{eq:our_param}) provides a better inductive bias for image DDPMs, leading to better generative models.
We use previously published, first-order UNet models pre-trained on various datasets (\Cref{tab:models}), add additional UNet heads to decode the various terms of our covariance model (details below), and optimize the parameters of these new heads w.r.t.\ the generalized NPR or OCM objectives. We systematically compare our approach with results previously reported for the same first-order models but with diagonal covariance approximations\footnote{Our code is built on previous work released by \citep{ou2024diffusion,bao2022estimating} for fair comparison, and can be found in \url{https://github.com/mtkresearch/highdiff}}.
Our results are summarized in \Cref{fig:both-better} and \Cref{tab:results}. 

\paragraph{Model structure and training} We use the same parameter sharing strategy as used in \citet{bao2022estimating, ou2024diffusion}, where the (pretrained) first-order noise predictor $\boldsymbol\epsilon_\theta$ and the covariance model $\mathcal{E}_\phi$ share most of their parameters, as follows:
\begin{equation}
    \label{eq:NN}
    \boldsymbol\epsilon_\theta(\bm{x}_t,t) = \mathrm{NN}_1(\mathrm{UNet}(\bm{x}_t,t;\theta_1), \theta_2), \quad \mathcal{E}_\phi(\bm{x}_t,t) = \mathrm{NN}_2(\mathrm{UNet}(\bm{x}_t,t;\theta_1), \phi)
\end{equation}
Here, $\theta_1$ and $\theta_2$ are \emph{fixed} parameters of the pretrained model, and $\text{NN}_2(\cdot; \phi)$ is a model that outputs the three key components of \Cref{eq:our_param} ($\boldsymbol\varepsilon_\phi, C_\phi, \bm{d}_\phi$) and which we train using the \emph{same} dataset and noising schedule as were used to pretrain the first-order model.
In more detail, $\boldsymbol\varepsilon_\phi$ receives input from the last \emph{up}-block layer in the UNet, while $C_\phi$ and $\bm{d}_\phi$ receive input from the last \emph{middle}-block layer. Indeed, we reasoned that the former might require pixel-level information, while the latter two might benefit from more abstract features.
Thus, our parameterization only requires training a smaller neural network  compared to the UNet.
Moreover, when compared with diagonal covariance models ($\boldsymbol\varepsilon_\phi$ only), our non-diagonal model only requires the addition of two smaller components ($C_\phi$ and $\bm{d}_\phi$) which adds negligible overhead.
Refer to \Cref{tab:our_architec} for detailed model architectures.

\begin{table}[b]
  \caption{Different covariance estimation methods ranked by increasing expressiveness}
  \label{tab:review}
  \centering
  \begin{tabular}{lll}
    \toprule
    Covariance &  Type   &  Intuition \\
    \midrule
    Heuristic large $\beta_t$ \citep{ho2020denoising} & Isotropic constant & Cov. of $q(\bm{x}_t|\bm{x}_{t-1})$ \\
    Heuristic small $\tilde{\beta}_t$ \citep{ho2020denoising} & Isotropic constant & Cov. of $q(\bm{x}_t|\bm{x}_{t-1},\bm{x}_0)$ \\
    SN-diagonal \citep{bao2022estimating} & Diagonal $\bm{x}_t$-dependent & Learn from data, $\mathbb{E}(\boldsymbol\epsilon_t^2|\bm{x}_t)$\\
    NPR-diagonal \citep{bao2022estimating} & Diagonal $\bm{x}_t$-dependent & Learn from data, Cov$(\boldsymbol\epsilon_t|\bm{x}_t)$\\
    OCM-diagonal \citep{ou2024diffusion} & Diagonal $\bm{x}_t$-dependent & Learn from score, $\nabla_{\bm{x}_t}\log\tilde{q}(\bm{x}_t,t) $\\
    LowRank \citep{meng2021higherorder} & Low-rank $\bm{x}_t$-dependent & Learn from data, Cov$(\boldsymbol\epsilon_t|\bm{x}_t)$ \\
    NPR-K-DCT (Ours) & Full $\bm{x}_t$-dependent & Learn from data, Cov$(\boldsymbol\epsilon_t|\bm{x}_t)$\\
    OCM-K-DCT (Ours) & Full $\bm{x}_t$-dependent & Learn from score, $\nabla_{\bm{x}_t}\log\tilde{q}(\bm{x}_t,t) $\\
    \bottomrule
  \end{tabular}
\end{table}
\paragraph{Datasets \& compared methods} Following the experimental setting of \citet{bao2022estimating}, we evaluate our full covariance model across several datasets and associated pre-trained first-order score networks: CIFAR10 \citep{krizhevsky2009cifar} with linear (LS; \citealp{ho2020denoising}) and cosine (CS; \citep{iddpm}) noising schedules, CelebA \citep{liu2015deep}, down-sampled ImageNet ($64 \times 64$; \citealp{deng2009imagenet}), and LSUN Bedroom \citep{yu2015lsun}. We borrow most of the implementation details and hyperparameter from \citet{bao2022estimating} and compare our results with those previously reported for constant, diagonal heuristic covariance and for $\bm{x}_t$-dependent diagonal covariance, summarized in \Cref{tab:review}. 

\begin{table}
  \caption{FID score and NLL across various datasets with different sampling steps. Colors denote $\textcolor{Bittersweet}{\textbf{1}^\textbf{st}}$ and $\textcolor{NavyBlue}{\textbf{2}^\textbf{nd}}$ best (i.e.\ lowest) FID and NLL values.}
  \label{tab:results}
  \centering
  \resizebox{\columnwidth}{!}{\input{tables/fid-nll}}
\end{table}

\begin{figure}
    \centering
    \includegraphics[width=\linewidth]{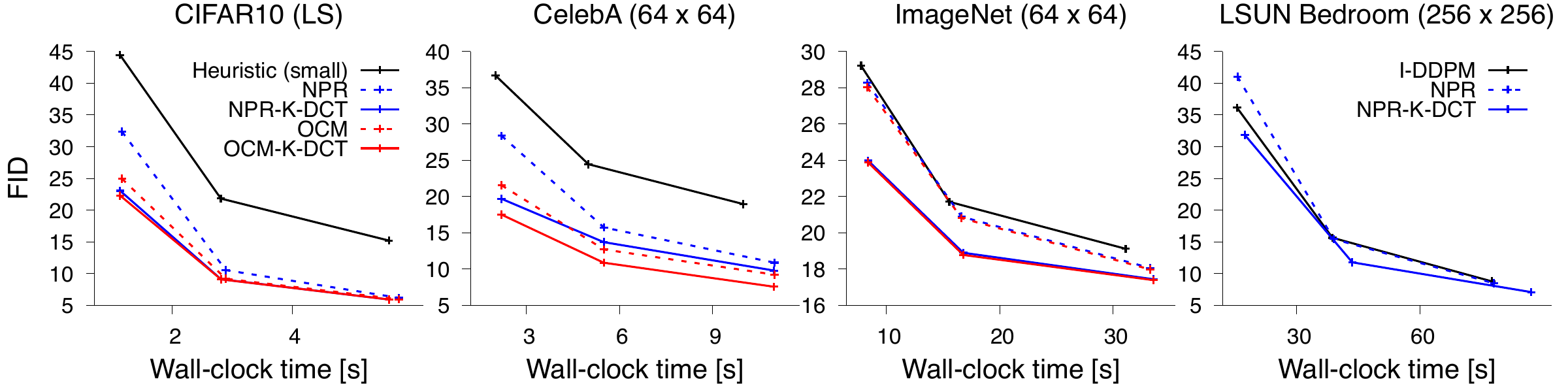}
    \caption{Wall-clock sampling time vs.\ FID for different datasets and covariance models (standard heuristic, diagonal, ours). The sampling time is an average measurement for a batch size of 128 on the CIFAR10, CelebA and ImageNet datasets, and 64 on the LSUN Bedroom dataset, all on a single A6000-48GB GPU. CIFAR10, CelebA and LSUN Bedroom are tested on $K=\{10,25,50\}$ steps, while ImageNet is tested on $K=\{25,50,100\}$ (hence three points per curve).}
    \label{fig:wall-clock}
\end{figure}

\paragraph{Evaluation}
For evaluation, we focus on two key metrics: statistical goodness of fit as measured by the average negative log-likelihood (NLL; $-\log p(\bm{x})$) of test images, and perceptual quality of generated images as measured by the FID \citep{heusel2017gans}.
We approximate the NLL via the standard evidence lower-bound \citep{ho2020denoising},
$-\log q(\bm{x}) \leq \mathbb{E}_q\log \frac{p(\bm{x}_{0:T:k})}{q(\bm{x}_{1:T:k}|\bm{x}_0)} \equiv -L_\text{ELBO}(\bm{x})$,
where $p(\cdot)$ denotes the Markov denoising process, and $k$ denotes the number of skipped steps.
The lower bound becomes tighter with more sampling steps, i.e. with smaller $k$.
Here we have conservatively evaluated all inverses and log determinants involved in the ELBO using direct Cholesky decompositions, but see \Cref{fig:suppl:slogdet} for an efficient stochastic estimator that is applicable at scale (\Cref{sec:detail_training}).
As shown in the lower block of \Cref{tab:results}, our K-DCT model consistently outperforms diagonal models in terms of NLL, regardless of the training objective used, be it score derivative-based (OCM) or MMSE-based (NPR).
As expected, improvements are more significant in the case of fewer sampling steps where a diagonal covariance no longer provides a good approximation.

These improvements in NLL are largely reflected in FID improvements (see upper block of \Cref{tab:results} where FIDs were evaluated based on 50k generated samples), especially in more aggressive skip-step regimes (lower $K$) and for larger images.
Although FID and NLL are known to be somewhat loosely related (e.g.\ the `squared-noise' (SN) diagonal approximation tends to do well in terms of FID but poorly on likelihoods; \citealp{bao2022estimating}), overall our K-DCT model exhibits the best tradeoff between these two evaluation metrics amongst all models (\Cref{fig:both-better}).
Example samples generated by our method can be found in supplementary materials. To quantify the additional compute overhead that K-DCT introduces, we present FID against wall-clock sampling time in \Cref{fig:wall-clock}, where we also include  experiments with the higher-resolution ($256\times 256$) LSUN Bedroom dataset.

Lastly, we investigate the benefits of directly learning the eigenvalues in frequency-domain comparing to using a low-rank structure. 
The model performance in FID along with the computation complexity are shown in in \Cref{sec:lowrank_results}, \Cref{tab:lowrank}. 
As expected, we find that our K-DCT model consistently outperforms ``diag+low-rank'' models, which in turn outperform purely diagonal models. 
With increasing rank (until $r=50$ ($4.8\%$ of max.) for CIFAR10 and $r=100$ ($2.4\%)$ for CelebA), the FID decreases but there is still a large gap between low-rank models and our full-rank K-DCT. 
To understand why, we examined the eigenvalue spectra of the posterior covariances at various points in the sampling process as shown in \Cref{sec:lowrank_results}, \Cref{fig:eigenspectrum}. 
These eigenvalues exhibit a power-law decay spanning several $(>3)$ orders of magnitude (especially mid-way through the denoising process), indicating that they cannot accurately be rank-truncated. 
Notice that the memory consumption of low-rank models increases significantly with higher ranks, while our model keeps negligible memory overhead. 

To investigate the harm of fixing to the DCT basis when the underlying data lacks translation invariance, (e.g. the CelebA dataset), we carried out some ablation studies by relaxing the spatial eigenbasis to learnable matrices (see details in \Cref{sec:ablation}). The improvements are only modest -- much smaller than the improvement made by the original K-DCT over a purely diagonal model (\Cref{tab:ablations}).

\section{Discussion \& Limitations} \label{sec:limit}
In summary, modeling important elements of natural image statistics can be done in an efficient way through our K-DCT parameterization, and improves image DDPMs especially in the regime of few denoising steps.
Given how strongly non-diagonal denoising posterior covariances are (\Cref{fig:illustration}C, top), and how much of this non-diagonal structure the K-DCT model appears to capture (\Cref{fig:illustration}C, bottom), it is perhaps surprising that the performance gains are not more striking; in particular, it is difficult to match FID results from distilled models (e.g. \citealp{zhou2024score}; on the other hand, these one-step models lack a tractable likelihood).   
Perhaps a fundamental limitation of the broader `covariance modeling approach' is that denoising posteriors in the skip-step regime are far from Gaussian, such that Gaussian sampling is not appropriate even with the right covariance.
This problem is analogous to that encountered in second-order optimization, whereby modeling the curvature of the loss function leads to larger, more aggressive parameter updates, but these can easily leave the `trust region' where the underlying quadratic approximation is valid.
In the same way that second-order optimization benefits strongly from adaptive damping \citep{martens2010deep}, second-order sampling might benefit from input-dependent adaptation of the step size (implicitly damping the posterior covariance).

Finally, while the generalizability of our method to non-image data remains an open question, it could potentially be applied to other domains where data exhibits approximate translation invariance, such as audio and speech (see \Cref{fig:suppl:audio} for a proof of principle on speech data). In fact, perhaps paradoxically, one of the datasets where our K-DCT model performed best (relative to diagonal models) is CelebA, i.e.\ images of faces that clearly lack translation invariance.
Thus, we speculate that K-DCT-like `full' covariance models that model non-diagonal elements even crudely may lead to performance gains even when the underlying data lacks symmetries.  

\paragraph{Acknowledgment} We thank the anonymous reviewers for their constructive feedback. This work was supported in part by EPSRC / MediaTek Research UK (iCase studentship to R. Xia).

\bibliography{main}
\bibliographystyle{tmlr}

\newpage
\appendix
\section{Appendix}

\subsection{Skip-step DDPM}
\label{sec:skipstep}
Similar to the affine transformation of \Cref{eq:affine,eq:affinecov} that relates the predicted mean and covariance of the noise to the 1-step image posterior, we can relate the same quantities to the \emph{skip-step} image posterior $q(\bm{x}_s|\bm{x}_t) \approx \mathcal{N}(\bm{x}_s; \boldsymbol\mu_s(\bm{x}_t), \Sigma_s(\bm{x}_t)) $ for $s<t$, as follows:
%
\begin{align}
        \boldsymbol\mu_s(\bm{x}_t,t; \theta) &=
        \frac1{\sqrt{\bar\alpha_{s:t}}}
        \left(\bm{x}_t - \frac{1-\bar\alpha_{s:t}}{\sqrt{1-\bar\alpha_t}}\boldsymbol\epsilon_\theta(\bm{x}_t,t)\right) \label{eq:kstep-mean}\\
        \Sigma_s(\bm{x}_t,t; \phi) &=
        \frac1{\bar\alpha_{s:t}(1-\bar\alpha_t)}
        \left((1-\bar\alpha_{s:t})(\bar\alpha_{s:t}-\bar\alpha_t)I + (1-\bar\alpha_{s:t})^2 \mathcal{E}_\phi(\bm{x}_t,t)\right) \label{eq:kstep-cov}
\end{align}
with the standard notation:
\begin{equation}
\bar\alpha_t \triangleq \prod_{t'=0}^t \alpha_{t'}
\qquad 
\bar\beta_t \triangleq 1 - \bar\alpha_t
\qquad 
\bar\alpha_{s:t} \triangleq \prod_{t'=s}^t(1-\beta_{t'})
\qquad
\bar\alpha_{s:t} = \frac{\bar\alpha_t}{\bar\alpha_s}
\end{equation}

\subsection{Expressiveness of the DCT parameterization} \label{sec:expressiveness}

\begin{figure}[h]
    \centering
    \includegraphics[width=0.5\linewidth]{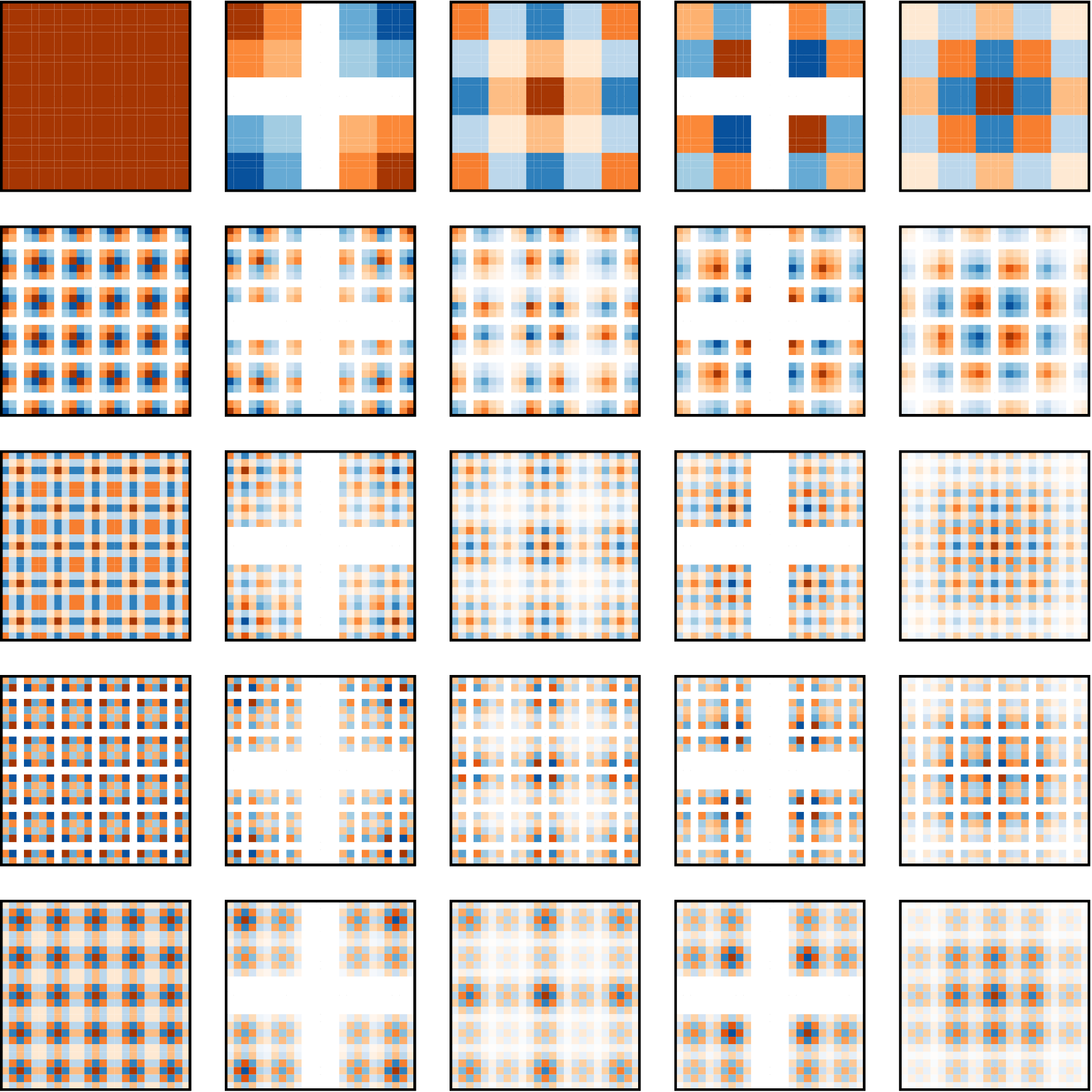}
    \caption{\textbf{Expressiveness of the DCT covariance parameterization}. Each panel $(i,j)$ shows $(F \otimes F)^\top \text{diag}(e_{i+dj}) (F \otimes F)$ where $F$ is the $d$-points DCT matrix, and $e_k$ is the $k^\text{th}$ row of the identity matrix $I_{d^2}$ (here, $d=5$). Thus, these are the primitive basis functions from which any $(F \otimes F)^\top \text{diag}(\bm\lambda) (F \otimes F)$ in \Cref{eq:our_param} can be assembled.}
    \label{fig:expressiveness}
\end{figure}

\Cref{fig:expressiveness} shows the basis functions from which the spatial (achromatic) component of \Cref{eq:our_param} is assembled parametrically.

\begin{figure}[h]
    \centering
    \includegraphics[width=\linewidth]{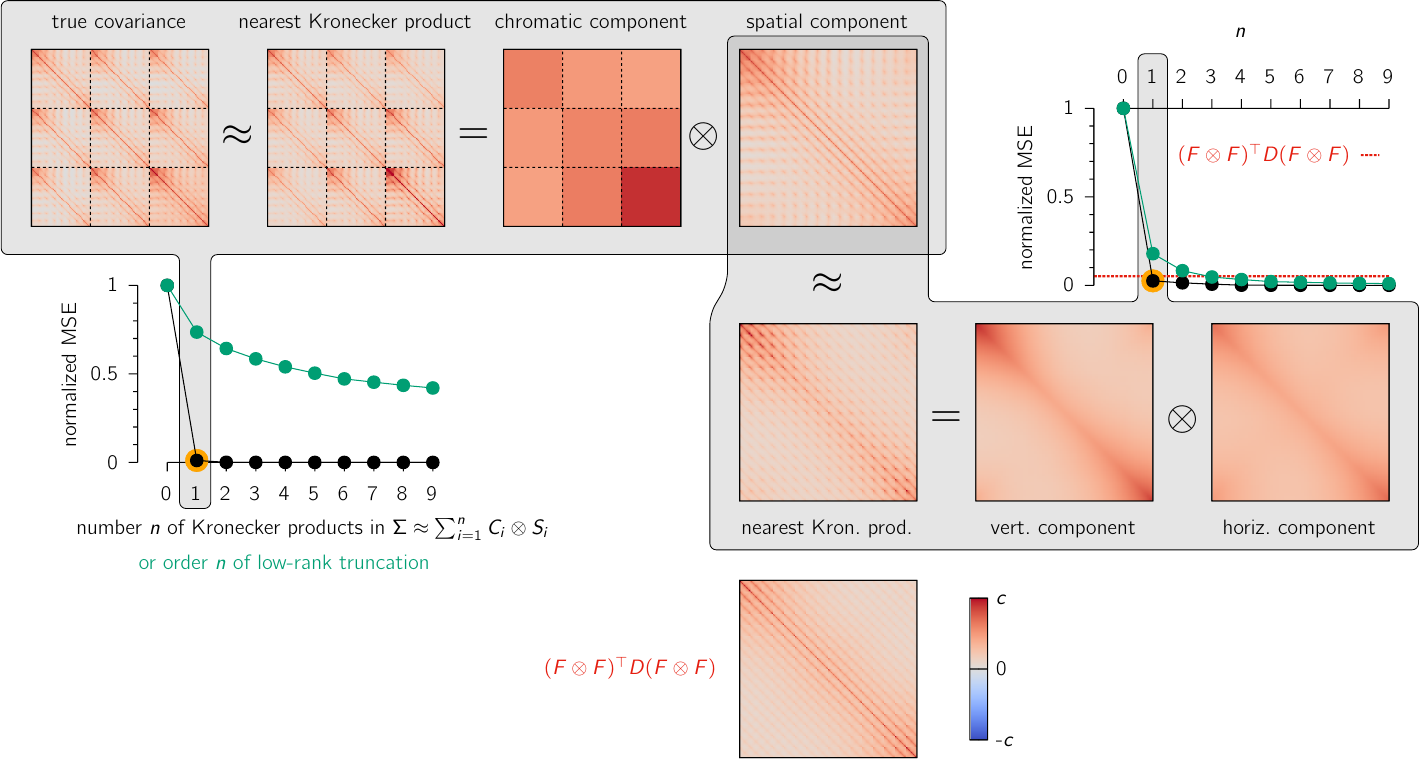}
    \caption{\textbf{Accuracy of our covariance parameterization for modeling the marginal (i.e.\ prior) ImageNet covariance}.
    \textbf{Left}:~normalized MSE (black) obtained when approximating the marginal ImageNet covariance $\Sigma$ by a sum of $n$ spatio-chromatic Kronecker products (“$\text{colors}(3\times 3) \otimes \text{pixels}(d^2 \times d^2)$”). The optimal decomposition of order $n$ has an analytical form \citep{van1993approximation}.
    Our approximation in \Cref{eq:our_param} corresponds to $n=1$, and is nearly perfect here; this best single Kronecker product approximation is shown at the top, along with the two corresponding factors. We will call the spatial factor $S \in \mathbb{R}^{d^2 \times d^2}$). 
    For comparison, we also show the SVD-based rank-$n$ truncation of $\Sigma$ (green).
    \textbf{Right:}~normalized MSE (black) obtained when approximating the spatial factor $S$ (c.f.\ above) with a sum of vertical-horizontal Kronecker products ($[d \times d] \otimes [d \times d]$).
    For $n=1$, this type of approximation is close to, but not exactly the same, as what we propose in \Cref{eq:our_param} for modeling the spatial component.
    It would be the same if (i) we constrained our $D\equiv \text{diag}(\bm\lambda)$ to itself be a Kronecker product of two smaller diagonals, but (ii) learned the eigenbasis instead of forcing it to be the DCT matrix $F$.
    Again, a single unconstrained Kronecker product provides an excellent approximation here, indicating that a Kronecker-structured eigenbasis is empirically justified.
    Moreover, fixing the eigenbasis to $F$ but still learning a full diagonal $D$ (our main proposal) does nearly as well (dashed red; $(F\otimes F)^\top D (F \otimes F)$) whilst having lower computational complexity.
    \textbf{In summary}, this figure shows that \textbf{the ImageNet dataset has approximately \emph{separable} spatio-chromatic components}, and is \textbf{sufficiently \emph{translation invariant}} for its spatial structure to be compactly described using the DCT.
    \label{fig:kronecker-imagenet}}
\end{figure}

\begin{figure}[h]
    \centering
    \includegraphics[width=\linewidth]{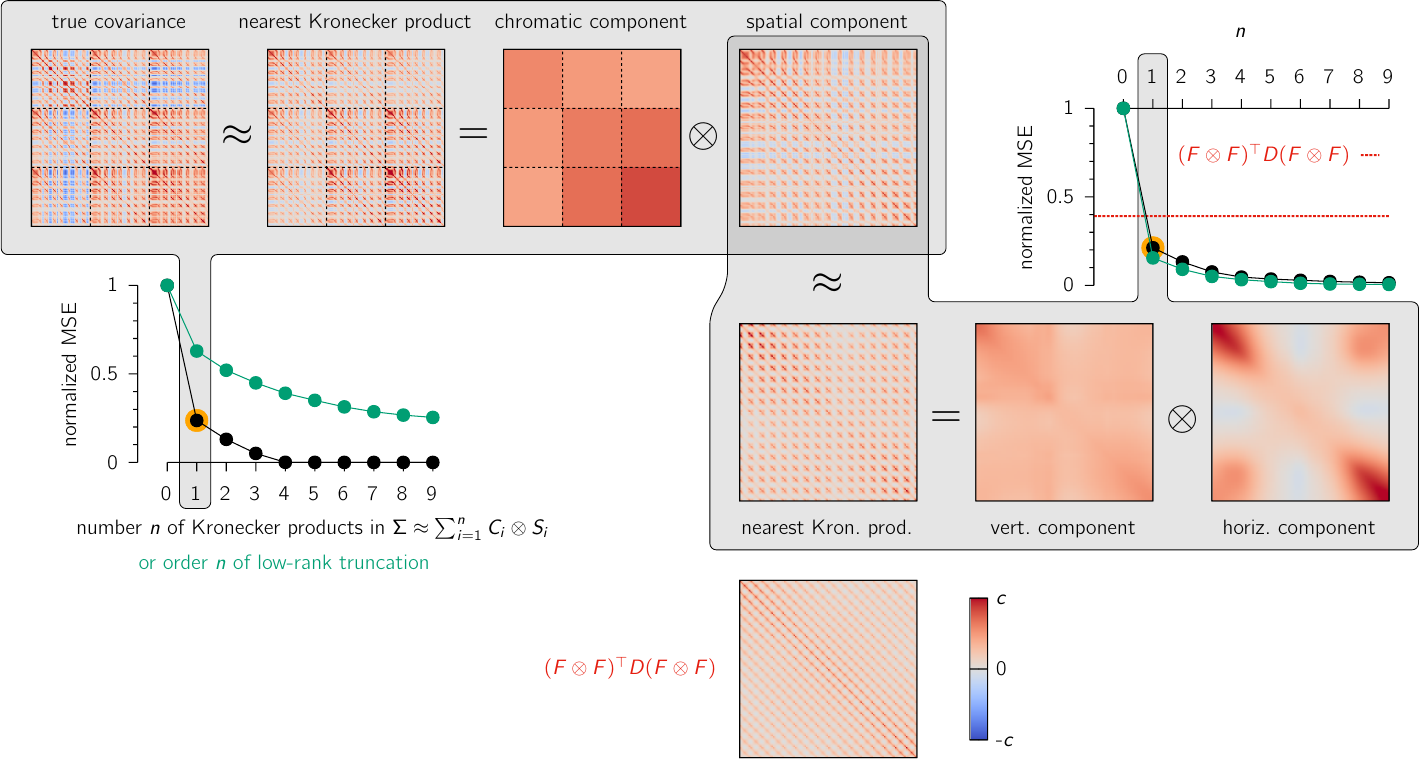}
    \caption{Same as \Cref{fig:kronecker-imagenet}, but for modelling the CelebA marginal (i.e.\ prior) covariance.
    Perhaps unsurprisingly, the spatio-chromatic structure of this particular dataset is not as separable as for ImageNet; this is likely due to certain locations in the image being dominated by certain colors (e.g.\ celebrities aren't known for their blue noses). It is also less translation invariant (i.e.\ less diagonalizable in the DCT eigenbasis) owing to the nature of these centered portraits.
    \label{fig:kronecker-celeba}}
\end{figure}

\begin{figure}[h]
    \centering
    \includegraphics[width=0.8\linewidth]{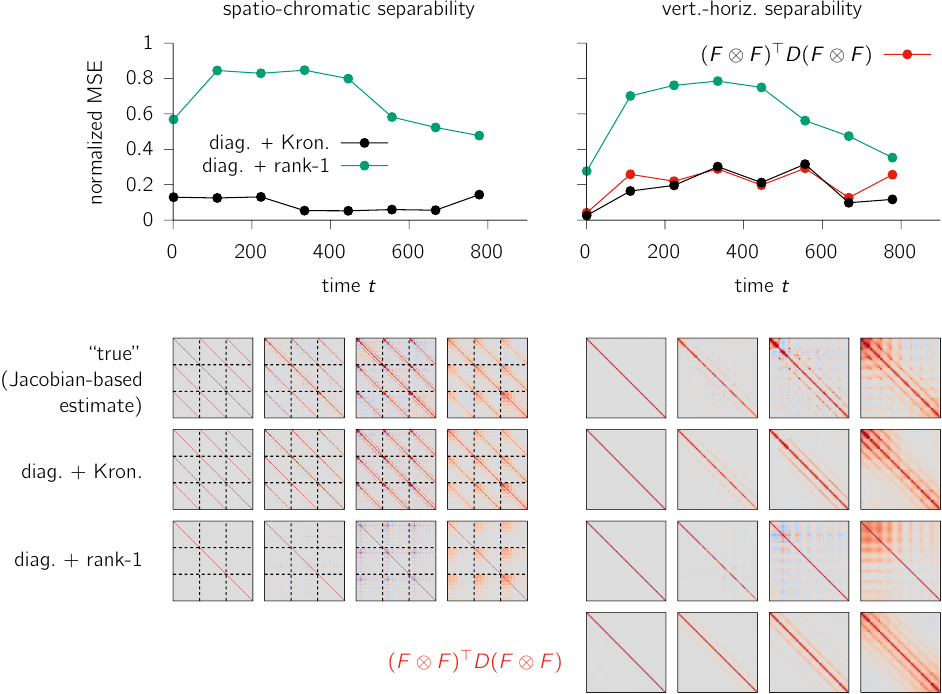}
    \caption{Accuracy of \Cref{eq:our_param} for modeling conditional covariances at different stages of denoising (time $t$) on CIFAR-10.
    \textbf{Left}:~at each time, we have a “true” covariance obtained by differentiating through the score network (c.f. \Cref{eq:tweedie_2}) shown in the first row of images. As in \Cref{fig:kronecker-imagenet,fig:kronecker-celeba}, for each of these matrices we can find its nearest “diagonal + spatio-chromatic Kronecker product” approximation (second row). This yields a fairly good reconstruction MSE (black), much better than the nearest “diagonal + rank-1” approximation (green; third row of images). Both approximations were obtained through optimization.
    \textbf{Right}:~at each time, we can then focus on the spatial component of the Kronecker product identified on the left, and approximate it in the same ways as we did in \Cref{fig:kronecker-imagenet} (right) with $n=1$. This is shown with the same color code.
    In summary, we see that at all times, the use of Kronecker products at both levels outperforms low-rank forms, and that reducing the complexity by restricting the spatial component to a DCT-diagonalized form does not hurt much.
   \label{fig:kronecker-cifar}}
\end{figure}

\clearpage
\subsection{Posterior sampling for inverse problems}
\label{sec: inverse_problems}

\begin{figure}[htbp]
    \centering
    
    \begin{subfigure}[b]{0.75\textwidth}
        \centering
        \includegraphics[width=\textwidth]{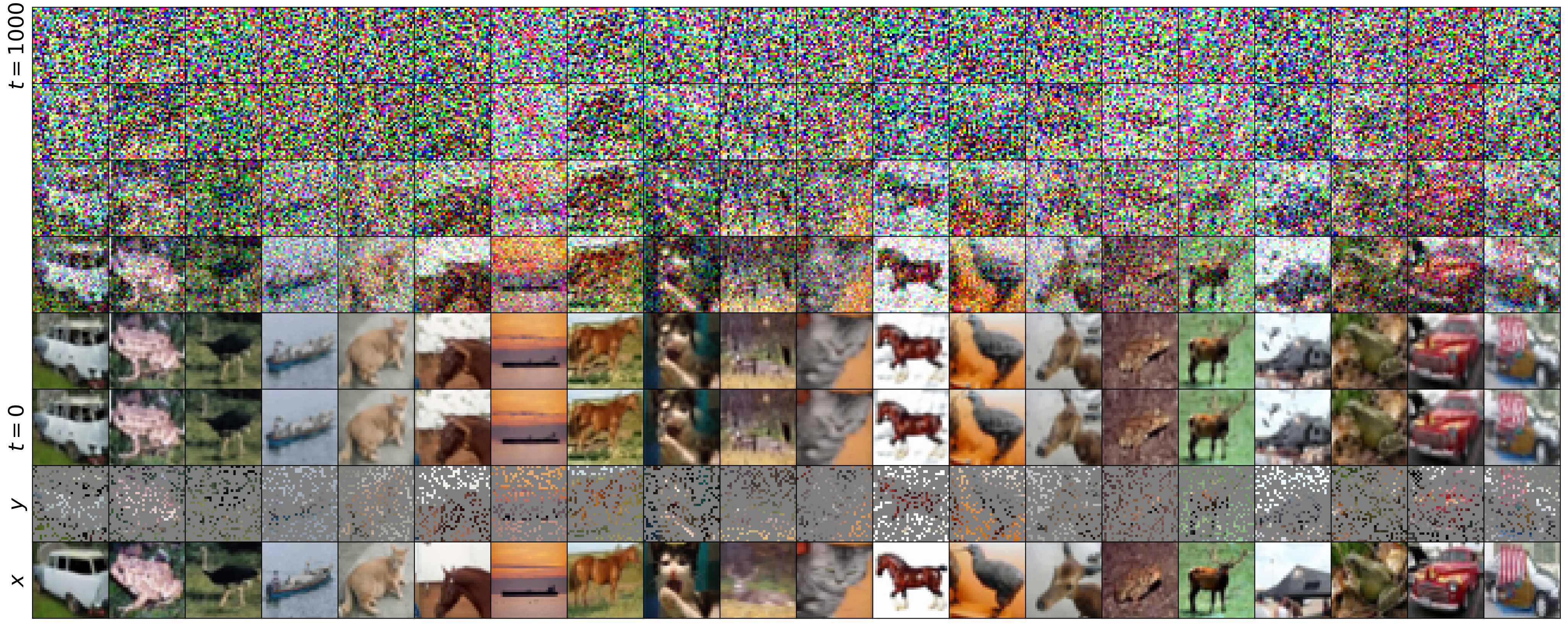}
        \caption{K-DCT ($K=10$ steps)}
    \end{subfigure}
    
    \begin{subfigure}[b]{0.75\textwidth}
        \centering
        \includegraphics[width=\textwidth]{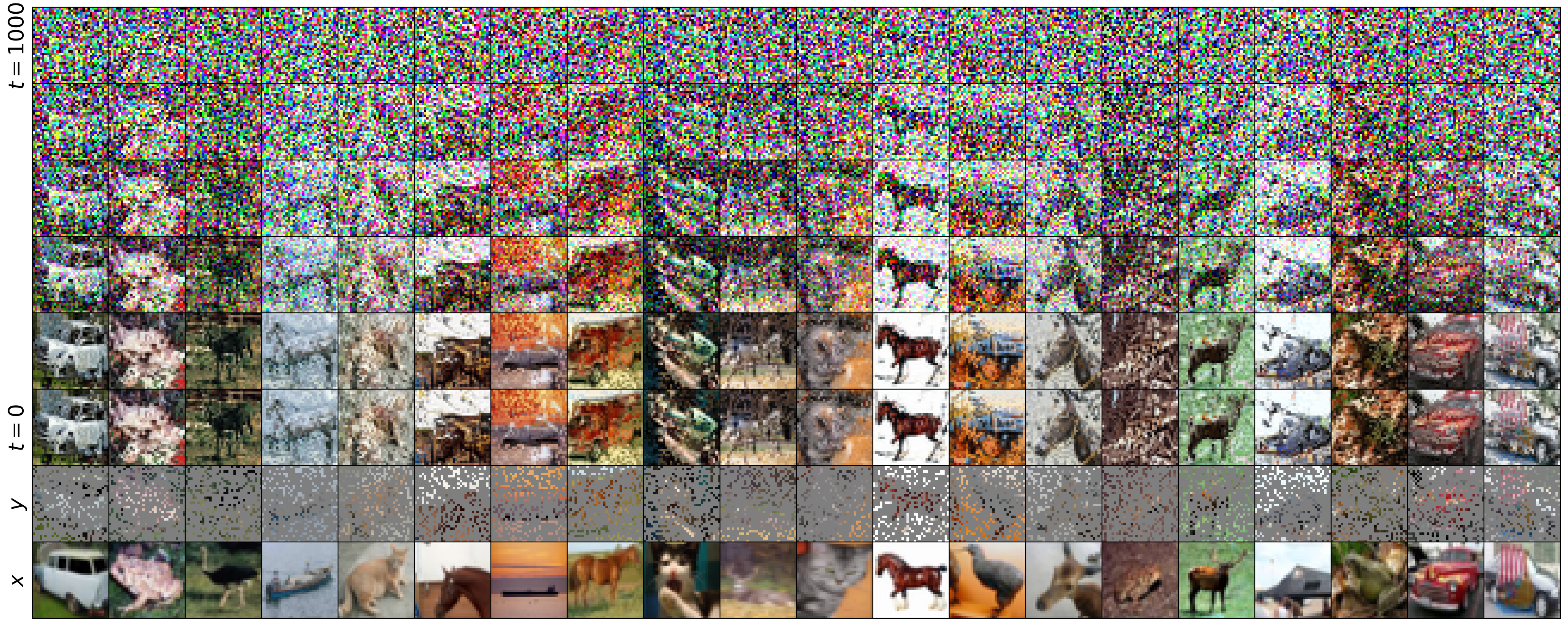}
        \caption{OCM ($K=10$ steps)}
    \end{subfigure}
    
    \begin{subfigure}[b]{0.75\textwidth}
        \centering
        \includegraphics[width=\textwidth]{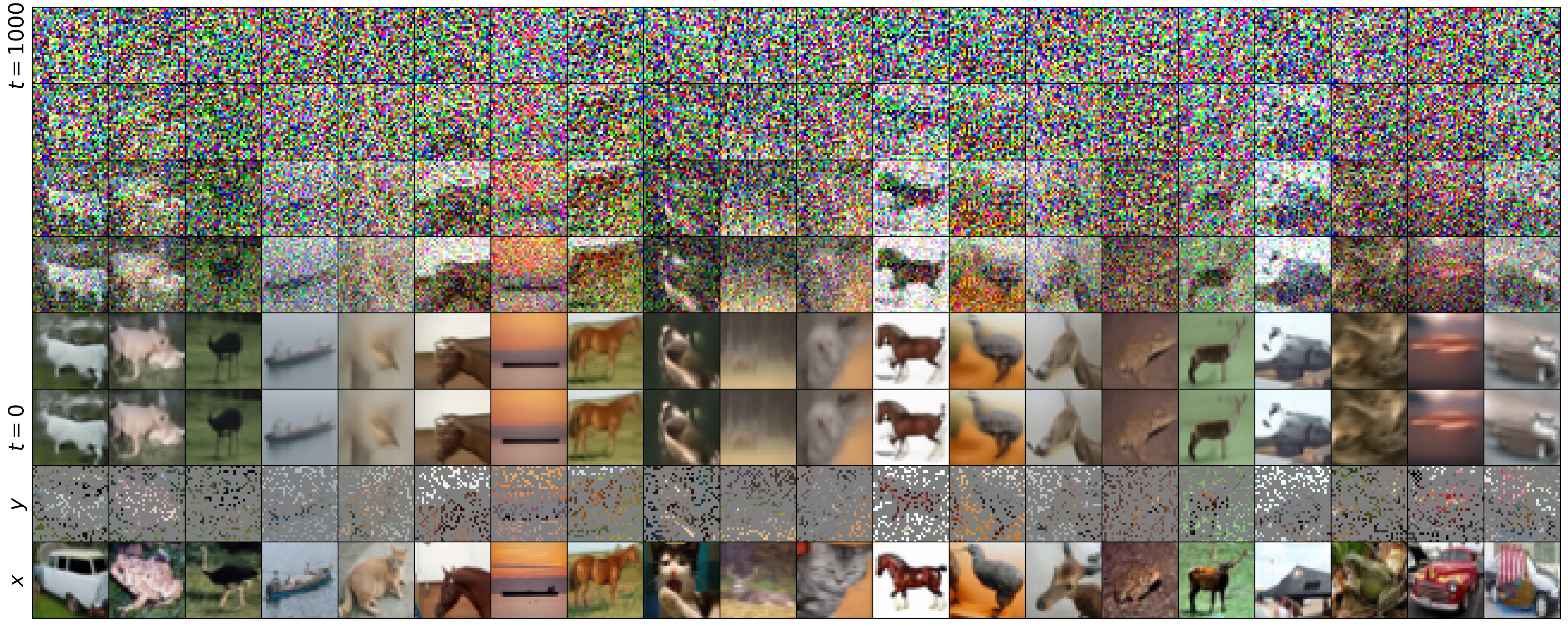}
        \caption{$\Pi$GDM ($K=10$ steps)}
    \end{subfigure}

    \begin{subfigure}[b]{0.75\textwidth}
        \centering
        \includegraphics[width=\textwidth]{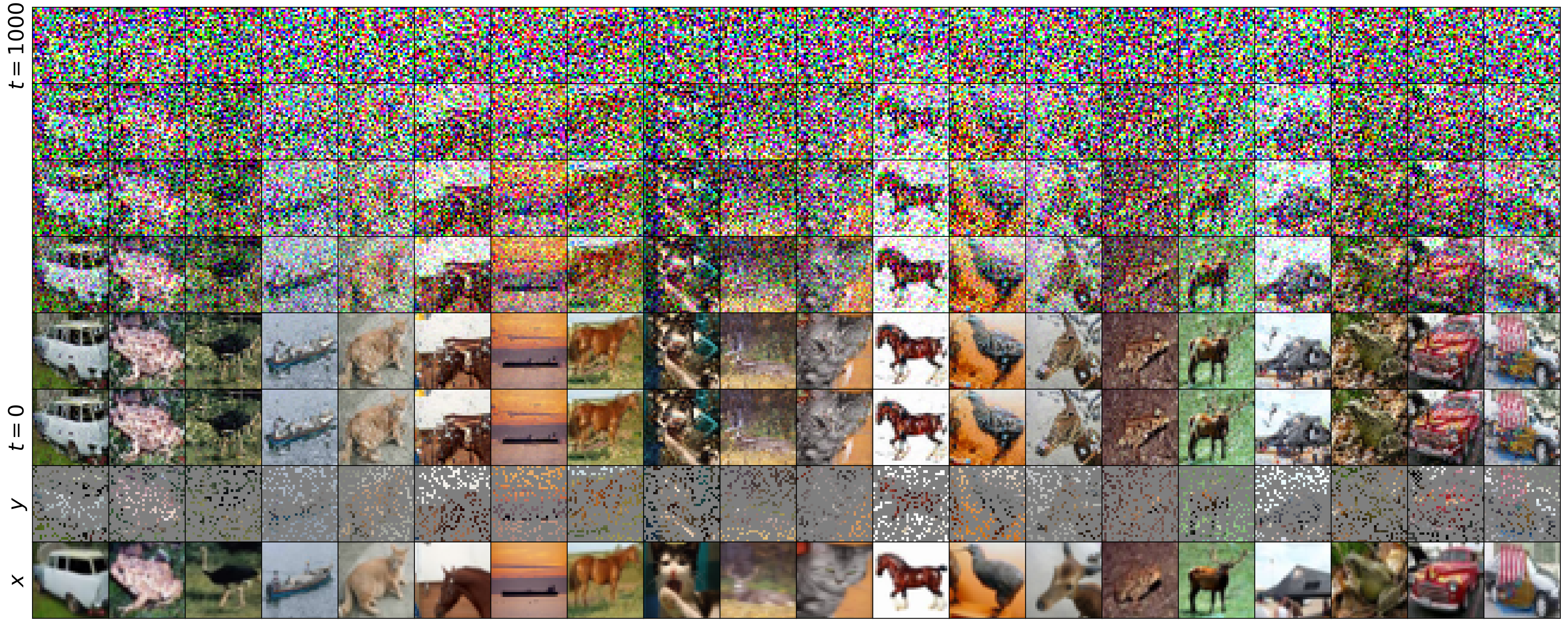}
        \caption{Tweedie's ($K=256$ steps)}
    \end{subfigure}
    
    \caption{Conditional posterior sampling for the noisy inpainting problem, with $\bm{y}=A\bm{x} + \sigma_{\bm{y}} \bm\epsilon$, where $A$ is a random mask that covers $75\%$ of the pixels, and additional Gaussian noise of $\sigma_{\bm{y}}=10^{-3}$. Each sub-figure shows conditional denoising sampling procedure for a different method but under the same seed, with the corrupted observations $\bm{y}$ and the original images $\bm{x}$ on the last two rows.} \label{fig:inverse}
\end{figure}

Our covariance models can be used out-of-the-box in conditional posterior sampling problems such as inpainting. 
An inverse problem assumes access to some measurements $\bm{y}$ or corrupted observations of the  original input $\bm{x}_0$. Here we focus on linear-Gaussian observations: 
$\bm{y} = A\bm{x}_0 + \bm\epsilon, \, \bm\epsilon \sim \mathcal{N}(0, \Sigma_{\bm{y}} = \sigma^2_{\bm{y}}\bm{I})$
where e.g.\ $A$ might perform a masking operation.
To sample from the posterior distribution $p(\bm{x}_0|\bm{y})$, a problem-specific score can be obtained via Bayes' rule as:
$\nabla_{\bm{x}_t}\log p_t(\bm{x}_t|\bm{y}) = \nabla_{\bm{x}_t} \log p_t(\bm{x}_t) + \nabla_{\bm{x}_t} \log p_t(\bm{y}|\bm{x}_t),$
where the first term is given by the unconditional score network, and the second term represents condition-specific guidance. 
As for unconditional sampling, we make a covariance-based Gaussian approximation, $q(\bm{x}_0|\bm{x}_t) = \mathcal{N}(\bm{x}_0; \mathbb{E}[\bm{x}_0|\bm{x}_t], \text{Cov}[\bm{x}_0|\bm{x}_t])$ (please refer to \Cref{eq:affine,eq:affinecov}). Given that the noising process and the observation model are both linear-Gaussian, one can estimate the likelihood score $\nabla_{\bm{x}_t} \log p_t(\bm{y}|\bm{x}_t)$ as \citep{song2023pseudoinverse, rozet2024learning},
\begin{equation}
    \nabla_{\bm{x}_t} \log p_t(\bm{y}|\bm{x}_t) = \nabla_{\bm{x}_t}\mathbb{E}[\bm{x}_0|\bm{x}_t]^\top A^\top (\Sigma_{\bm{y}} + A \text{Cov}[\bm{x}_0|\bm{x}_t]A^\top)^{-1}(\bm{y} - A\mathbb{E}[\bm{x}_0|\bm{x}_t]). \label{eq:guidance}
\end{equation}
When taking smaller steps, using a simple heuristics for $\text{Cov}[\bm{x}_0|\bm{x}_t]$ can generate samples with good quality. 
However, larger steps would require accurate modeling of the covariance and our K-DCT model comes to help. 
Specifically, our model provides estimates for both $\text{Cov}[\bm{x}_0|\bm{x}_t]$ and $\nabla_{\bm{x}_t}\mathbb{E}[\bm{x}_0|\bm{x}_t]^\top$ in the guidance term:
\makeatletter
\newcommand{\subalign}[1]{%
  \vcenter{%
    \Let@ \restore@math@cr \default@tag
    \baselineskip\fontdimen10 \scriptfont\tw@
    \advance\baselineskip\fontdimen12 \scriptfont\tw@
    \lineskip\thr@@\fontdimen8 \scriptfont\thr@@
    \lineskiplimit\lineskip
    \ialign{\hfil$\m@th\scriptstyle##$&$\m@th\scriptstyle{}##$\hfil\crcr
      #1\crcr
    }%
  }%
}

\begin{equation}
    \underbrace{\nabla_{\bm{x}_t}\mathbb{E}[\bm{x}_0|\bm{x}_t]^\top}_{
    \subalign{
        &\text{[K-DCT]} = \mathcal{E}_\phi(\bm{x}_t)/\sqrt{\bar\alpha_t}\\
        &\text{[$\Pi$GDM]} = \text{through VJP}\\
        &\text{[Tweedie's]} = \text{through VJP}
    }
    } A^\top (\Sigma_{\bm{y}} + A\underbrace{\text{Cov}[\bm{x}_0|\bm{x}_t]}_{
        \subalign{
        &\text{[K-DCT]} = \bar\beta_t\mathcal{E}_\phi(\bm{x}_t)/\bar\alpha_t\\
        &\text{[$\Pi$GDM]} = \bar\beta_t \bm{I}\\
        &\text{[Tweedie's]} = \text{through VJP}
    }
    }A^\top)^{-1}(\bm{y} - A\mathbb{E}[\bm{x}_0|\bm{x}_t]).
\end{equation}
We compare our method with diagonal covariance modeling OCM \citep{ou2024diffusion}, $\Pi$GDM that uses a heuristic covariance \citep{song2023pseudoinverse}, and a direct evaluation of Tweedie's formula \citep{rozet2024learning} using vector-Jacobian products. We consider a challenging denoising + inpainting painting problem, where $A$ masks out $75\%$ of the pixels uniformly and randomly, with additional \emph{i.i.d.}\ Gaussian noise ($\sigma_{\bm{y}}=10^{-3}$) on the CIFAR10 dataset. Quantitative results are shown in \Cref{tab:inverse} and qualitative samples are shown in \Cref{fig:inverse}. Our model has significantly better FID and classification accuracy (AC) when using $10$ steps sampling, where all models are tested with the same classification model (pretrained ResNet20).

\begin{table}[h]
    \centering
    \begin{tabular}{lccc|ccc}
        \toprule
        Method & Step & FID$\downarrow$ & AC$\uparrow$ & Step & FID$\downarrow$ & AC$\uparrow$  \\
        \midrule
        \rowcolor{gray!10} K-DCT (ours) & \multirow{4}{*}{10} & \textbf{14.28} & \textbf{72.00}\% & \multirow{4}{*}{100} & 5.12 & 82.42\% \\
        $\Pi$GDM \citep{song2023pseudoinverse} & & 33.89 & 47.48\% & & \textbf{4.01} & \textbf{84.08}\% \\
        OCM \citep{ou2024diffusion} & & 77.60 & 36.57\% & & 25.93 & 70.95\% \\
        Tweedie's \citep{rozet2024learning} & & - & - &  & 75.21 & 55.16\%\\
        \bottomrule
    \end{tabular}
    \caption{Inpainting+denoising results. FID and AC are reported on $50k$ samples.}
    \label{tab:inverse}
\end{table}

Notice that \cite{rozet2024learning} did not evaluate the vector-Jacobian products on the pretrained unconditional denoiser model, but the improved posterior sampling scheme they proposed can be used to test unconditional diffusion models. Due to numerical errors and imperfect pre-training, the Jacobian matrix is not guaranteed to be perfectly symmetric positive definite, hence our reproduction is far from being superior and we only report for $100$ sampling steps. In addition, the inverse operation in \Cref{eq:guidance} is suggested to be approximated by conjugate gradients (CG) since the covariance matrix should be symmetric positive definite. However, we find CG to perform poorly on our model and we currently choose to compute exact matrix inverse which can be time and memory consuming. Proposing a more efficient approximation method for the inverse is left for future work.

\subsection{Learning the covariance of the noise}
\label{sec:covofnoiseproof}
\paragraph{Lemma 1}(Use Tweedie's formula to derive the covariance of the noise) Given the DDPM noising process, $q(\bm{x}_t|\bm{x}_0)=\mathcal{N}(\bm{x}_t; \sqrt{\bar\alpha_t}\bm{x}_0 + \bar\beta_t I)$, we have that the covariance of the effective noise equals 
\begin{equation}
\label{eq:cov_noise_proof}
    \text{Cov}[\boldsymbol\epsilon_t|\bm{x}_t] =
    I - \sqrt{\bar\beta_t}\nabla_{\bm{x}_t}\boldsymbol\epsilon_\theta(\bm{x}_t,t)
\end{equation}

\emph{Proof.} Let the perfect noise predictor be $\boldsymbol\epsilon_\theta(\bm{x}_t,t) = \mathbb{E}\left[\frac{\bm{x}_t - \sqrt{\bar\alpha_t}{\bm{x}_0}}{\sqrt{\bar\beta_t}}|\bm{x}_t\right]$ which relates to the first order score through $s_1(\bm{x}_t) \equiv \nabla_{\bm{x}_t}\log\tilde{q}(\bm{x}_t) = -\frac{1}{\sqrt{\bar\beta_t}}\boldsymbol\epsilon_\theta(\bm{x}_t,t)$. Let the second-order score be $s_2(\bm{x}_t) \equiv \nabla_{\bm{x}_t}^2\log\tilde{q}(\bm{x}_t)$. From Tweedie’s first and second order formulae, we have the following:

\begin{align}
    \mathbb{E}[\bm{x}_0|\bm{x}_t] &= \frac{\bm{x}_t + \bar\beta_t s_1(\bm{x}_t)}{\sqrt{\bar\alpha_t}}\\
    \text{Cov}[\bm{x}_0|\bm{x}_t] &= \frac{\bar\beta_t}{\bar\alpha_t}\bigg(I + (1-\bar\alpha_t) s_2(\bm{x}_t)\bigg)
\end{align}

From the above equations, one can easily derive,
\begin{equation}
    \label{eq:tweedie_1_and_2}
    \mathbb{E}\left. \left[\bm{x}_0\bm{x}_0^\top - \frac{1}{\sqrt{\bar\alpha_t}}(\bm{x}_0\bm{x}_t^\top + \bm{x}_t\bm{x}_0^\top)\right|\bm{x}_t\right] = -\frac{1}{\bar\alpha_t}\bm{x}_t\bm{x}_t^\top + \frac{\bar\beta_t^2}{\bar\alpha_t}\bigg(s_1(\bm{x}_t)s_1(\bm{x}_t)^\top + s_2(\bm{x}_t)\bigg) + \frac{\bar\beta_t}{\bar\alpha_t}I
\end{equation}
Hence, we can solve for the covariance of the added noise through the following derivation:
\begin{equation}
    \begin{split}
        &\text{Cov}\left[\frac{\bm{x}_t-\sqrt{\bar\alpha_t}\bm{x}_0}{\sqrt{\bar\beta_t}}|\bm{x}_t\right] \\
        &= \mathbb{E}\left[\frac{(\bm{x}_t-\sqrt{\bar\alpha_t}\bm{x}_0)(\bm{x}_t-\sqrt{\bar\alpha_t}\bm{x}_0)^\top}{\bar\beta_t}\bigg|\bm{x}_t\right] - \mathbb{E}\left[\frac{\bm{x}_t - \sqrt{\bar\alpha_t}{\bm{x}_0}}{\sqrt{\bar\beta_t}}\bigg|x_t\right]\mathbb{E}\left[\frac{\bm{x}_t - \sqrt{\bar\alpha_t}{\bm{x}_0}}{\sqrt{\bar\beta_t}}\bigg|\bm{x}_t\right]^\top\\
        &= \frac{1}{\bar\beta_t}\bigg(
            \bm{x}_t\bm{x}_t^\top + \bar\alpha_t\mathbb{E}\bigg[\bm{x}_0\bm{x}_0^\top - \frac{1}{\sqrt{\bar\alpha_t}}(\bm{x}_0\bm{x}_t^\top + \bm{x}_t\bm{x}_0^\top)|\bm{x}_t\bigg]\bigg) -\epsilon_\theta(\bm{x}_t,t)\epsilon_\theta(\bm{x}_t,t)^\top\\
        &= \frac{1}{\bar\beta_t}\bigg(
            \bm{x}_t\bm{x}_t^\top + \bar\alpha_t\bigg(
                -\frac{1}{\bar\alpha_t}\bm{x}_t\bm{x}_t^\top + \frac{\bar\beta_t^2}{\bar\alpha_t}\bigg(s_1(\bm{x}_t)s_1(\bm{x}_t)^\top + s_2(\bm{x}_t)\bigg) + \frac{\bar\beta_t}{\bar\alpha_t}I
            \bigg)\bigg) -\boldsymbol\epsilon_\theta(\bm{x}_t,t)\boldsymbol\epsilon_\theta(\bm{x}_t,t)^\top\\
        &= \frac{1}{\bar\beta_t}\bigg(
            \bar\beta_t\epsilon_\theta(\bm{x}_t,t)\epsilon_\theta(\bm{x}_t,t)^\top + \bar\beta_t^2 s_2(\bm{x}_t) + \bar\beta_tI
            \bigg) -\boldsymbol\epsilon_\theta(\bm{x}_t,t)\boldsymbol\epsilon_\theta(\bm{x}_t,t)^\top\\
        &= I + \bar\beta_t s_2(\bm{x}_t) \\
        &= I - \sqrt{\bar\beta_t}\nabla_{\bm{x}_t}\boldsymbol\epsilon_\theta(\bm{x}_t,t)
    \end{split}
\end{equation}

As an alternative to using derivatives
of the noise predictor as in \Cref{eq:cov_noise_proof} to estimate the covariance of the noise, one can also obtain it as a least-squares estimator in the MMSE approach (c.f.\ \hyperref[sec:background]{Background}).
The derivation of this estimator begins with
\Cref{eq:tweedie_1_and_2} above, which shows that its r.h.s.\ is the minimizer of the following mean squared error:
\begin{equation}
    \begin{split}
    &\mathbb{E}_{q(\bm{x}_0)q(\bm{x}_t|\bm{x}_0)}\left\|    
        \left(\bm{x}_0 - \frac{1}{\sqrt{\bar\alpha_t}}\bm{x}_t\right)\left(\bm{x}_0 - \frac{1}{\sqrt{\bar\alpha_t}}\bm{x}_t\right)^\top - \frac{\bar\beta_t^2}{\bar\alpha_t}\bigg(s_1(\bm{x}_t)s_1(\bm{x}_t)^\top + s_2(\bm{x}_t)\bigg) - \frac{\bar\beta_t}{\bar\alpha_t}I
    \right\|^2_\text{F}\\
    &= \mathbb{E}_{q(\bm{x}_0)q(\bm{x}_t|\bm{x}_0)}\left\|    
        \frac{\bar\beta_t}{\bar\alpha_t}\left(\boldsymbol\epsilon_t\boldsymbol\epsilon_t^\top-I\right) - \frac{\bar\beta_t^2}{\bar\alpha_t}\left(s_1(\bm{x}_t)s_1(\bm{x}_t)^\top + s_2(\bm{x}_t)\right)\right\|^2_\text{F}\\
    &= \mathbb{E}_{q(\bm{x}_0)q(\bm{x}_t|\bm{x}_0)}\left\|
        \frac{\bar\beta_t}{\bar\alpha_t}\left(\boldsymbol\epsilon_t\boldsymbol\epsilon_t^\top-I\right) - \frac{\bar\beta_t^2}{\bar\alpha_t}\left(\frac{\boldsymbol\epsilon_\theta(\bm{x}_t,t)\boldsymbol\epsilon_\theta(\bm{x}_t,t)^\top}{1-\bar\alpha_t} + s_2(\bm{x}_t)\right)
        \right\|^2_\text{F}\\
    &= \mathbb{E}_{q(\bm{x}_0)q(\bm{x}_t|\bm{x}_0)}\left\|    
        \frac{\bar\beta_t}{\bar\alpha_t}\bigg(\boldsymbol\epsilon_t\boldsymbol\epsilon_t^\top-\boldsymbol\epsilon_\theta(\bm{x}_t,t)\boldsymbol\epsilon_\theta(\bm{x}_t,t)^\top-\left(I+\bar\beta_t s_2(\bm{x}_t)\right) \bigg)
    \right\|^2_\text{F} \\
     &= \mathbb{E}_{q(\bm{x}_0)q(\bm{x}_t|\bm{x}_0)}\left\|    
        \frac{\bar\beta_t}{\bar\alpha_t}\bigg(\boldsymbol\epsilon_t\boldsymbol\epsilon_t^\top-\boldsymbol\epsilon_\theta(\bm{x}_t,t)\boldsymbol\epsilon_\theta(\bm{x}_t,t)^\top- \text{Cov}(\boldsymbol{\epsilon}_t | \bm{x}_t) \bigg)
    \right\|^2_\text{F}
    \end{split}
\end{equation}
This derivation shows that the covariance of the noise is the MMSE estimator of $\boldsymbol\epsilon_t\boldsymbol\epsilon_t^\top-\boldsymbol\epsilon_\theta(\bm{x}_t,t)\boldsymbol\epsilon_\theta(\bm{x}_t,t)^\top$, which leads to the generalized NPR objective below.

\subsection{Efficient Training \& Sampling} \label{sec:eff_training_sampling}

\subsubsection{Efficient training: log-linear complexity loss}
When equipped with our parameterization of the posterior covariance in \Cref{eq:our_param}, one can evaluate the loss in \Cref{eq:npr_objective} and \Cref{eq:ocm_objective} in near-linear (in spatial resolution) complexity. Firstly, one can easily extend the objectives for the diagonal case to full covariance. In particular, for NPR, \Cref{eq:npr_objective} can be generalized and simplified as follow, (dependency on $(\bm{x}_t,t)$ is dropped for brevity and the 2D-DCT $F\otimes F$ is also shortened to simply $F$),
\allowdisplaybreaks

\begin{alignat}{2} 
    \mathcal{L}_{\text{NPR}}(\phi)   &= \mathbb{E}_{q_{\text{data}}(\bm{x}_0)q(\bm{x}_t|\bm{x}_0)} \mathrlap{\left[ \left\lVert
                             \mathcal{E}_\phi- (\boldsymbol\epsilon_t\boldsymbol\epsilon_t^\top - \boldsymbol\epsilon_\theta\boldsymbol\epsilon_\theta^\top) \right\rVert^2_\text{F} \right]}
                             \label{eq:npr_full}\\
                            &= \mathbb{E}_{q_{\text{data}}(\bm{x}_0) q(\bm{x}_t|\bm{x}_0)} \bigg[ \Big(
                                &\lVert & \mathcal{E}_\phi \rVert^2_\text{F} +(\boldsymbol\epsilon_t^\top \boldsymbol\epsilon_t)^2 + (\boldsymbol\epsilon_\theta^\top\boldsymbol\epsilon_\theta)^2 - 2(\boldsymbol\epsilon_t^\top\boldsymbol\epsilon_\theta)^2 \nonumber \\
                            &   &- &2 \boldsymbol\epsilon_t^\top \mathcal{E}_\phi \boldsymbol\epsilon_t + 2 \boldsymbol\epsilon_\theta^\top \mathcal{E}_\phi\boldsymbol\epsilon_\theta \Big) \bigg]\\
                            &= \mathbb{E}_{q_{\text{data}}(\bm{x}_0) q(\bm{x}_t|\bm{x}_0)} \bigg[ \Big(
                                &\lVert & \text{diag}(\boldsymbol\varepsilon_\phi) + \left(C_\phi C_\phi^\top \otimes F^\top D_\phi F\right) \rVert^2_\text{F} \nonumber \\
                            &               &- &2 \boldsymbol\epsilon_t^\top \left( \text{diag}(\boldsymbol\varepsilon_\phi) + C_\phi C_\phi^\top \otimes F^\top D_\phi F\right) \boldsymbol\epsilon_t \nonumber\\
                            &               &+ &2 \boldsymbol\epsilon_\theta^\top \left( \text{diag}(\boldsymbol\varepsilon_\phi) + C_\phi C_\phi^\top \otimes F^\top D_\phi F \right) \boldsymbol\epsilon_\theta \\
                            &               &+ & \text{ const. w.r.t. } \phi \nonumber \Big) \bigg]\\
                            &= \mathbb{E}_{q_{\text{data}}(\bm{x}_0) q(\bm{x}_t|\bm{x}_0)} \bigg[ \Big(
                                            &\lVert & \boldsymbol\varepsilon_\phi \rVert^2_\text{F} + \lVert C_\phi C_\phi^\top \rVert^2_\text{F} \cdot \lVert D_\phi\rVert^2_\text{F} \nonumber \\
                            &               &+ &2\text{Tr}\left(\text{diag}(\boldsymbol\varepsilon_\phi)  \left(C_\phi C_\phi^\top \otimes F^\top D_\phi F\right)\right) \label{eq:trace_term} \\
                            &               &- &2 \boldsymbol\epsilon_t^\top \left(C_\phi C_\phi^\top \otimes F^\top D_\phi F\right) \boldsymbol\epsilon_t - 2 \boldsymbol\varepsilon_\phi^\top (\boldsymbol\epsilon_t\odot\boldsymbol\epsilon_t) \label{eq:mvp1} \\
                            &               &+ &2 \boldsymbol\epsilon_\theta^\top \left(C_\phi C_\phi^\top \otimes F^\top D_\phi F \right) \boldsymbol\epsilon_\theta + 2 \boldsymbol\varepsilon_\phi^\top(\boldsymbol\epsilon_\theta\odot\boldsymbol\epsilon_\theta) \label{eq:mvp2} \\
                            &               &+ & \text{ const. w.r.t. } \phi \nonumber \Big)\bigg]
\end{alignat}
where the trace term (\Cref{eq:trace_term}) can be efficiently calculated as follow, (denoting $i$ as pixel in 3D, $c$ as color channel, and $k$ as pixel in 2D),
\begin{align}
    \text{Tr}\left(\text{diag}(\boldsymbol\varepsilon_\phi) \left(C_\phi C_\phi^\top \otimes F^\top D_\phi F\right)\right) &= \sum_i (\boldsymbol\varepsilon_{\phi})_i \left(C_\phi C_\phi^\top \otimes F^\top D_\phi F\right)_{ii}\\
    &= \sum_c\sum_{k} (\boldsymbol\varepsilon_{\phi})_{ck} (C_\phi C_\phi^\top)_{cc} (F^\top D_\phi F)_{kk}\\
    &= \sum_c (C_\phi C_\phi^\top)_{cc} \sum_{k}(\boldsymbol\varepsilon_{\phi})_{ck} \sum_{k'} F^\top_{kk'} (d_{\phi}) _{k'} F_{k'k}\\
    &= \sum_c (C_\phi C_\phi^\top)_{cc} \sum_{k}(\boldsymbol\varepsilon_{\phi})_{ck} \sum_{k'} F_{kk'}^{\top\odot2} (d_{\phi}) _{k'} \\
    &= \sum_c (C_\phi C_\phi^\top)_{cc} \left((\boldsymbol\varepsilon_{\phi})_c^\top F^{\top\odot2} d_\phi\right)
\end{align}
where $[\cdot]^{\odot2}$ means element-wise square. The above shows that only matrix-vector products and squared Frobenius norm are required for optimizing the objectives. When calculating the matrix-vector product of $\left(C_\phi C_\phi^\top \otimes F^\top D_\phi F\right)\text{vec}(V)$ in \Cref{eq:mvp1} and \Cref{eq:mvp2} for $V=\boldsymbol\epsilon_t$ or $V=\boldsymbol\epsilon_\theta(\bm{x}_t,t)$, it can be effectively calculated using FFT as shown in \Cref{sec:kdct}. However, in practice, we find matrix-multiplication to work most efficiently for GPU. The full algorithm of training using the NPR objectives is shown in \Cref{alg:npr_training}.

\begin{algorithm}
    \caption{Computing the NPR objective for our parametrization as in \Cref{eq:npr_full}}
    \label{alg:npr_training}
    \begin{algorithmic}[1]
       \Require Covariance model components $\{\boldsymbol\varepsilon_\phi, C_\phi, \bm{d}_\phi\}$, pretrained first-order (noise predictor) model $\boldsymbol\epsilon_\theta$, a batch of samples $(\bm{x}_0, t)$ and a noise scheduler for computing $\alpha_t, \beta_t$.
       \Ensure $g$ as a batch estimate of \Cref{eq:npr_full}
       \State Compute the noised samples, $\bm{x}_t \gets \sqrt{\bar\alpha_t} \bm{x}_0 + \sqrt{\bar\beta_t} \boldsymbol\epsilon_t$ with $\boldsymbol\epsilon_t \sim \mathcal{N}(0,I)$
       \State Compute model outputs $\boldsymbol\varepsilon \gets \boldsymbol\varepsilon_\phi(\bm{x}_t, t)$, $CC^\top \gets C_\phi(\bm{x}_t, t) C_\phi(\bm{x}_t, t)^\top$ and $\bm{d} \gets \bm{d}_\phi(\bm{x}_t, t)$
       \State Compute $\text{norm} \gets \lVert\boldsymbol\varepsilon \rVert^2_\text{F} + \lVert CC^\top\rVert^2_\text{F} \cdot \lVert \bm{d}\rVert^2_\text{F} + 2\sum_c (CC^\top)_{cc} \left< \boldsymbol\varepsilon_c, \text{2D-iDCT}^{\odot2}(\bm{d})\right>$ \\ \text{\# linear-logarithmic, $\boldsymbol\varepsilon_c$ is the $c$-th channel}
       \State Define $f(\bm{v}) = \left<\bm{v}, \text{2D-iDCT}(\bm{d}\star \text{2D-DCT}(CC^\top\bm{v})) + \boldsymbol\varepsilon \odot \bm{v} \right>$ \\
       \text{\# linear-logarithmic, $\star$ is a broadcasting product}
       \State Compute $\text{trace} \gets f(\boldsymbol\epsilon_t) - f(\boldsymbol\epsilon_\theta(\bm{x}_t,t))$
       \State Compute $g \gets \text{norm} - 2\cdot\text{trace}$
    \end{algorithmic}
\end{algorithm}

For OCM, \Cref{eq:ocm_objective} can be generalized to full covariance (\Cref{eq:cov_noise_proof}) and approximated using the Hutchinson's trick \cite{hutchinson1989stochastic} as follow, ( $\bm{v}\sim p(\bm{v})$ is a Rademacher random variable with entries $\pm 1$, dependency on $(\bm{x}_t,t)$ is dropped for brevity and the 2D-DCT $F\otimes F$ is shortened to simply $F$),

\begin{align} 
    \mathcal{L}_{\text{OCM}}(\phi)&= \mathbb{E}_{q_{\text{data}}(\bm{x}_0)q(\bm{x}_t|\bm{x}_0)}\left[ \left\lVert
                             \mathcal{E}_\phi- (I - \sqrt{\bar\beta_t}\nabla_{\bm{x}_t}\boldsymbol\epsilon_\theta) \right\rVert^2_\text{F} \right] \label{eq:ocm_full}\\
                            &= \mathbb{E}_{q_{\text{data}}(\bm{x}_0) q(\bm{x}_t|\bm{x}_0)} \Bigg[
                                \lVert  \mathcal{E}_\phi \rVert^2_\text{F} - 2\text{Tr}\left(\mathcal{E}_\phi\right)
                            + 2\sqrt{\bar\beta_t}\text{Tr}(\mathcal{E}_\phi\nabla_{\bm{x}_t}\boldsymbol\epsilon_\theta\big) + \text{const. w.r.t. }\phi\Bigg] \\
                            &\approx \mathbb{E}_{q_{\text{data}}(\bm{x}_0) q(\bm{x}_t|\bm{x}_0)q(\bm{v})} \Bigg[
                             \bm{v}^\top\mathcal{E}_\phi^\top \mathcal{E}_\phi\bm{v} - 2\bm{v}^\top \mathcal{E}_\phi\bm{v}
                            +  2\sqrt{\bar\beta_t}\bm{v}^\top\mathcal{E}_\phi^\top\underbrace{\nabla_{\bm{x}_t}\boldsymbol\epsilon_\theta\bm{v}}_{\text{JVP}} + \text{ const. w.r.t. }\phi\Bigg]
\end{align}
where $\mathcal{E}_\phi(\bm{x}_t,t)\bm{v}$ can be efficiently evaluated  similarly as mentioned when calculating $\mathcal{L}_{\text{NPR}}$, and $\nabla_{\bm{x}_t}\boldsymbol\epsilon_\theta(\bm{x}_t,t)\bm{v}$ can be efficiently evaluated using forward-mode AD that does not depend on $\phi$. The full algorithm of training using the OCM objective is shown in \Cref{alg:ocm_training}.

\begin{algorithm}
    \caption{Computing the OCM objective for our parametrization as in \Cref{eq:ocm_full}}
    \label{alg:ocm_training}
    \begin{algorithmic}[1]
       \Require Covariance model components $\{\boldsymbol\varepsilon_\phi, C_\phi, \bm{d}_\phi\}$, pretrained first-order (noise predictor) model $\boldsymbol\epsilon_\theta$, a batch of samples $(\bm{x}_0, t)$ and a noise scheduler for computing $\alpha_t, \beta_t$.
       \Ensure $g$ as a batch estimate of \Cref{eq:ocm_full}
       \State Sample two random variable, $\boldsymbol\epsilon_t \sim \mathcal{N}(0,I)$, $\bm{v}\sim \text{Bernoulli}(0.5) * 2-1$ \text{\ \ \# Rademacher samples}
       \State Compute the noised samples, $\bm{x}_t \gets \sqrt{\bar\alpha_t} \bm{x}_0 + \sqrt{\bar\beta_t} \boldsymbol\epsilon_t$
       \State Compute model outputs $\boldsymbol\varepsilon \gets \boldsymbol\varepsilon_\phi(\bm{x}_t, t)$, $CC^\top \gets C_\phi(\bm{x}_t, t) C_\phi(\bm{x}_t, t)^\top$ and $\bm{d} \gets \bm{d}_\phi(\bm{x}_t, t)$
       \State Compute $\mathcal{E}\bm{v} \gets \text{2D-iDCT}(\bm{d}\star \text{2D-DCT}(CC^\top\bm{v})) + \boldsymbol\varepsilon \odot \bm{v}$ \\
       \text{\ \ \ \# linear-logarithmic, $\star$ is a broadcasting product}
       \State Compute $H\bm{v} \gets \text{JVP}(\boldsymbol\epsilon_\theta(\cdot,t), \text{primals}=\bm{x}_t, \text{tangents}=\bm{v})$ \text{\ \ \ \# stop-gradients}
       \State Compute $g \gets \left< \mathcal{E}\bm{v}, \mathcal{E}\bm{v} \right> - 2 \left<\bm{v}, \mathcal{E}\bm{v} \right> + 2 \sqrt{\bar\beta_t} \left<\mathcal{E}\bm{v}, H\bm{v}\right>$
    \end{algorithmic}
\end{algorithm}

\subsection{Details of experiments} \label{sec:exp_details}
In this section, we provide detailed experimental setup for \Cref{sec:exp}, including details for model architectures, training, inference and evaluation. Our setups largely follow those used by \citep{bao2022estimating,ou2024diffusion} for fair comparison. 
\paragraph{Details of pretrained first-order model} We have used the same group of pretrained models as in \citet{bao2022estimating} and \citet{ou2024diffusion}. \Cref{tab:models} lists the pretrained models and noise schedulers used in our experiments. These models are effectively noise predictors which relates to the first-order score as $\boldsymbol\epsilon_\theta(\bm{x}_t,t) = -\sqrt{\bar\beta_t}s_1(\bm{x}_t)$. 
\begin{table}[h]
    \centering
    \caption{Pretrained first-order model used in our experiments, and details of the noise schedulers}
    \begin{tabular}{lccc}
        \toprule
         Datasets & Noise scheduler & Sampling steps & Pretrained model\\
         \midrule
         CIFAR10 & Linear & $1000$ & \citet{bao2022estimating} \\
         CIFAR10 & Cosine & $1000$ & \citet{bao2022estimating} \\
         CelebA $64 \times 64$&  Linear & $1000$ & \citet{song2020score} \\
         ImageNet $64 \times 64$& Cosine & $4000$ & \citet{nichol2021improved} \\
         LSUN Bedroom $256 \times 256$& Cosine & $1000$ & \citet{nichol2021improved}\\
         \bottomrule
    \end{tabular}
    \label{tab:models}
\end{table}
\paragraph{Details of K-DCT second-order model} For fair comparison, we follow most of the parameterization as per \citep{bao2022estimating,ou2024diffusion} for all models. The architecture details of $\text{NN}_1$ and $\text{NN}_2$ (including three components $\{\boldsymbol\varepsilon_\phi, C_\phi, \bm{d}_\phi\}$) in \Cref{eq:NN} are provided in \Cref{tab:our_architec}, where $\text{Conv}$ denotes the convolutional layer, $\text{Res}$ denotes the residual block for dependence on time $t$, and $\text{MLP}$ denotes multi-layer perceptron layers with 1-2 hidden layers. $[\cdot]_\text{mid}$ and $[\cdot]_\text{last}$ denote positions of the heads, whether they receive output of the UNet from the \emph{middle}-block layer or the \emph{last} up-block layer.
\begin{table}[h]
    \centering
    \caption{Architecture details of parametric heads for the first and second-order model as in \Cref{eq:NN}}
    \begin{tabular}{lcccc}
         \toprule
         Datasets & $\text{NN}_1$ & $\text{NN}_2,\boldsymbol\varepsilon_\phi$ & $\text{NN}_2,C_\phi$ & $\text{NN}_2,\bm{d}_\phi$\\
         \midrule
         CIFAR10 (LS) & $\text{Conv}_\text{last}$ & $\text{Conv}_\text{last}$ & $\text{MLP}_\text{mid}$ & $\text{MLP}_\text{mid}$ \\
         CIFAR10 (CS) & $\text{Conv}_\text{last}$ & $\text{Conv}_\text{last}$ & $\text{MLP}_\text{mid}$ & $\text{MLP}_\text{mid}$ \\
         CelebA $64 \times 64$ & $\text{Conv}_\text{last}$ & $ \text{Conv}_\text{last}$ & $(\text{Res} + \text{MLP})_\text{mid}$ & $(\text{Res} + \text{MLP})_\text{mid}$ \\
         ImageNet $64 \times 64$ & $\text{Conv}_\text{last}$ & $ (\text{Res} + \text{Conv})_\text{last}$ & $(\text{Res} + \text{MLP})_\text{mid}$ & $(\text{Res} + \text{MLP})_\text{mid}$ \\
         LSUN Bedroom $256 \times 256$ & $\text{Conv}_\text{last}$ & $ (\text{Res} + \text{Conv})_\text{last}$ & $(\text{Res} + \text{MLP})_\text{mid}$ & $(\text{Res} + \text{MLP})_\text{mid}$ \\
         \bottomrule
    \end{tabular}
    \label{tab:our_architec}
\end{table}

\paragraph{Cost of training and inference time} In \Cref{tab:cost}, we provide empirical comparisons of cost of time for model function evaluation, between the diagonal second-order model and our parameterization of the full covariance model. For training, we provide the average time of one iteration of training update for batch size of $128$, which includes evaluation of the corresponding objective, followed by backpropagation. For sampling, we provide the average time of one step of denoising, i.e. calculating $\boldsymbol\mu_s(\bm{x}_t) + \Sigma_s^{1/2}(\bm{x}_t)\boldsymbol\xi$ for $\bm{x}_t$ of batch size $128$. It is clear that our K-DCT model has a negligible additional time compared to diagonal covariances both for training and sampling.
As for model memory, the additional memory cost of MLPs for the two additional components, $\{C_\phi, \bm{d}_\phi\}$, is much smaller than the original UNet.

\begin{table}
    \centering
    \caption{Averaged time (in millisecond) of one iteration of update w.r.t different objectives (training) and one step of denoising (sampling) for a batch size of $128$ on an A6000-48GB GPU}\label{tab:cost}
    \begin{tabular}{lcccccc}
         \toprule
          & \multicolumn{4}{c}{Training} & \multicolumn{2}{c}{Sampling}\\
          \cmidrule(r){2-5} \cmidrule(r){6-7}
          & \multicolumn{2}{c}{NPR} & \multicolumn{2}{c}{OCM} & & \\
         \cmidrule(r){2-3}  \cmidrule(r){4-5} \cmidrule(r){6-7}
         Datasets & diagonal & K-DCT & diagonal & K-DCT & diagonal & K-DCT \\
         \midrule
         CIFAR10 & 118.74 & 124.82 (+ 5.1\%) & 310.02 & 318.89 (+ 2.8\%) & 115.74 & 112.49\\
         CelebA $64 \times 64$ & 285.25 &297.37 (+ 4.2\%) & 680.49 &727.64 (+ 6.9\%)  & 219.71 & 219.47 \\
         ImageNet $64 \times 64$ & 230.34 & 238.16 (+ 3.4\%) & 581.51 & 656.89 (+ 11.4\%) & 332.61 & 335.63 \\
         \bottomrule
    \end{tabular}
\end{table}

\subsection{Details of training, inference and evaluation} \label{sec:detail_training}
\paragraph{Training details} We use the AdamW optimizer with a learning rate of $1\times 10^{-4}$ and train for $500K$ iterations across all datasets. The batch size is $128$ for all datasets and we select the checkpoint saved every $10K$ iterations with the best FID on $1K$ generated samples with full sampling steps. We train our models using one A6000-48GB GPU for CIFAR10, CelebA, ImageNet; and evaluate on the same machine but one B200-180GB GPU for ImageNet.

\paragraph{Sampling details} As all previous papers have noticed, covariance clipping is as crucial to performance as mean clipping in diffusion models. Covariance clipping is trivial in the diagonal case, because at the penultimate sampling step, the condition $\|\Sigma_1(\bm{x}_2)\|_\infty\mathbb{E}(\boldsymbol\epsilon) \leq \frac{2}{255}y$ can be enforced through element-wise clipping on the diagonal. However, for our K-DCT model this would require forming the full covariance matrix . To circumvent this, we propose two methods. The first is applying scaling on both side of the covariance, $S^{1/2}\Sigma_1(\bm{x}_2)S^{1/2}$, where 
\begin{equation}
    S^{cij}=\begin{cases}
    1, & \text{if \ \ diag}(\Sigma_1)^{cij} \leq e \\
    e/\text{diag}(\Sigma_1)^{cij}, & \text{if \ \ diag}(\Sigma_1)^{cij} > e
    \end{cases}
\end{equation}
and $e$ is the corresponding threshold. The above only requires access to the diagonal part of the predicted covariance, which is easy and fast to evaluate using the K-DCT structure. The second way is directly clipping the generated sample, $\tilde{\boldsymbol\xi}=\Sigma_1(\bm{x}_2)\boldsymbol\xi$, element-wisely by $(-|e\boldsymbol\xi|, |e\boldsymbol\xi|)$. We find that the second method gives better results. We use the same value of $y$ as in \citet{bao2022estimating}.

\paragraph{Evaluation details} All results are evaluated on the exponential moving average of the trained models with a rate of $0.9999$. For computing the evidence lower-bound, we calculate the following terms over the entire test set,
\begin{align}
     -L_\text{ELBO}(\bm{x}) &= \mathbb{E}_\text{noising process} \!\bigg[\text{KL}(q(\bm{x}_T|\bm{x}_0)\|p(\bm{x}_T)) \nonumber \\
     &+ \sum_{t\in \{k+1:T:k\}}
     \text{KL}(q(\bm{x}_{t-k}|\bm{x}_t,\bm{x}_0)\|p_{\theta,\phi}(\bm{x}_{t-k}|\bm{x}_t))-\log p_\theta(\bm{x}_0|\bm{x}_1)\bigg].
   \label{eq:elbo_estimation}
\end{align}
The last sampling step $p(\bm{x}_0|\bm{x}_1)$ is approximated by likelihood of discrete image data, which is the same receipt used by \cite{bao2022estimating, ho2020denoising}. Notice that, to calculate the KL terms in \Cref{eq:elbo_estimation}, one needs to calculate the (log)-determinant and inverse of $\mathcal{E}_\phi$ which our parameterization does not provide an efficient way of evaluation. The results shown in \Cref{tab:results} are calculated by forming explicitly the $3D\times 3D$ full matrix which is time consuming but given this is necessary neither for training nor for sampling, we left this for future research. As for computing the FID score, $50K$ samples are generated, whose distribution is compared with the reference distribution statistics using the full training set for CIFAR10 and ImageNet, and $50K$ training samples for CelebA (published by \citealp{baoanalytic}). Results in \Cref{tab:results} are obtained using the same random seed as in \citep{bao2022estimating,ou2024diffusion}. Additionally, variance across three different random seeds is shown in \Cref{tab:errorbar}.

\begin{table}[]
    \centering
    \caption{Mean and standard deviation of FID and NLL.}
  \label{tab:errorbar}
  \centering
  \resizebox{\columnwidth}{!}{\input{tables/std-mean.tex}}
\end{table}

\begin{figure}[!h]
    \centering
    \includegraphics[width=\linewidth]{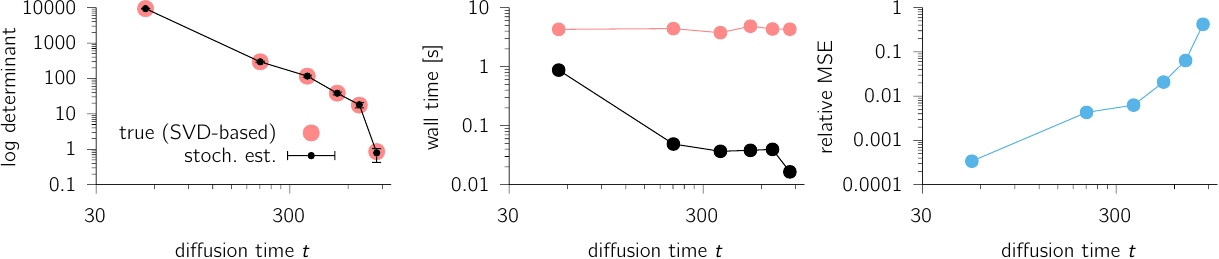}
    \caption{Scalable, efficient evaluation of the log determinant for log-likelihood computations.
    \textbf{Left}:~Stochastic estimation (black; mean and quartiles from 100 independent runs) vs.\ exact   value (red) of the log determinant involved in the log-likelihood at different stages of denoising ($t$) on CIFAR10.
    Here, stochastic estimation is done as shown in \Cref{eq:slogdettrick} with a \emph{single} probe vector $\bm\xi$.
    \textbf{Center}:~Wall time spent in computing the log determinant (CPU).
    \textbf{Right}:~Relative MSE in log-determinant estimation, estimated from 100 independent runs. Note that the terms that dominate the log-likelihood (small $t$) are also those that are approximated the best.  For large $t$, the noise in Hutchinson's trace estimator with only one probe vector is comparable to the log-det itself, hence the high relative MSE. However, these terms contribute very little to the overall NLL, and they are also the cheapest to compute, so one could use many more probe vectors if accuracy was really required there.
    }
    \label{fig:suppl:slogdet}
\end{figure}

Although for evaluation purposes we computed all log-determinants involved in the KL terms of \Cref{eq:elbo_estimation} by the direct (Cholesky) method, we note that our covariance parameterization affords fast matrix-vector products, and that this property could be used for evaluating log-determinants more efficiently in high-dimension (higher-resolution images).
We provide a proof of principle in \Cref{fig:suppl:slogdet}, using a stochastic estimator the log-determinant based on adaptive numerical quadrature and Hutchinson's trace estimator \citep{rutten2020non}. This estimator is based on the following identity: 
\begin{equation}
   \log | \Sigma | = \text{Tr}\left[ \log\Sigma \right] = \left\langle \bm\xi^\top 
    (\log \Sigma) \bm\xi \right\rangle_{\bm\xi}
\end{equation}
where the expectation is over any spherical distribution $p(\bm\xi)$.
To compute $(\log\Sigma)\bm\xi$ products, we rely on the integral representation of the matrix logarithm:
\begin{equation}
   (\log \Sigma)\bm\xi = \int_0^1 ds (\Sigma - I)
    \left[ s \Sigma + (1-s)I \right]^{-1} \bm\xi
    \label{eq:slogdettrick}
\end{equation}
For a fixed $\bm\xi$, this can be computed by any numerical quadrature algorithm, using conjugate gradients (CG) to compute the integrand at any $s$ -- CG iterations make good use of efficient matrix-vector products.
Note that CG at some $s$ can be warm-started by the solution already obtained at another, nearby $s'$.
Note also that in early stages of denoising, $\Sigma_t$ can be very ill-conditioned, requiring finer time steps for accurately approximating the integral in \Cref{eq:slogdettrick}.
We therefore use an adaptive, Gauss-Kronrod solver, which spends more time near $s=1$.
Finally, we remark that an important property of $\log\Sigma$ is that its eigenvalues have much less spread than those of $\Sigma$ itself. This means that Hutchinson's trace estimator is very accurate even with a single probe vector $\bm\xi$ (\Cref{fig:suppl:slogdet}).

\subsection{More results}\label{sec:lowrank_results}
\begin{figure}
    \centering
    \begin{subfigure}{\textwidth}
        \includegraphics[width=\textwidth]{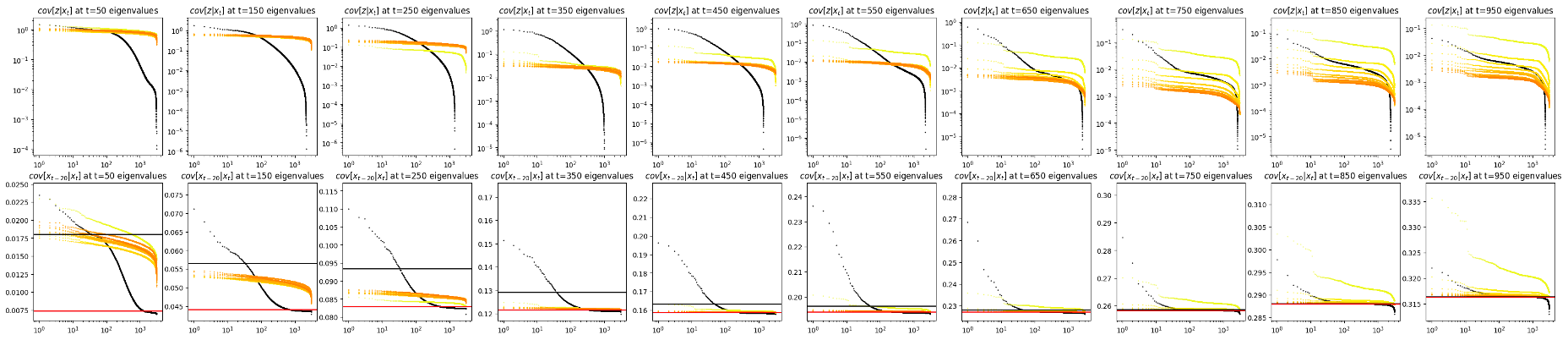}
        \subcaption{Diagonal Covariance}
    \end{subfigure}
    
    \begin{subfigure}{\textwidth}
        \includegraphics[width=\textwidth]{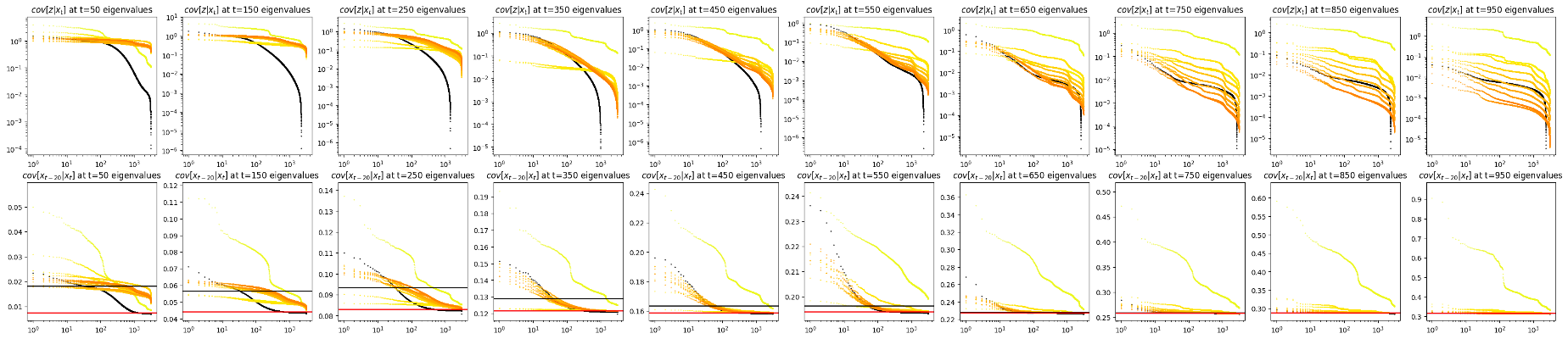}
        \subcaption{K-DCT Covariance}
    \end{subfigure}
    \caption{Eigen-spectrum of the posterior covariance at various points in the sampling process. Columns from left to right correspond to time from $t=50$ to $t=T-50$. Eigenvalues of the \textbf{true} posterior covariance are shown in \textbf{black} (obtained by evaluating the derivative of the pretrained score network at some random sample as in \Cref{eq:tweedie_2}). Eigenvalues of the fitted covariance with (a) diagonal assumption or (b) our proposed K-DCT structure are shown in orange, where the color from yellow to orange shows the dynamics along training iterations. Within each sub-figure, the first row shows eigenvalues of covariance of the noise, Cov$[\boldsymbol\epsilon_t|\bm{x}_t]$, while the second row shows eigenvalues of the skip-step covariance used during sampling, Cov$[\bm{x}_{t-k}|\bm{x}_t]$, for $k=20$. Horizontal lines in red and black indicate the value of `large', $\beta_t$, and `small', $\tilde{\beta}_t$, heuristics, respectively.}
    \label{fig:eigenspectrum}
\end{figure}

We carried out comparisons with low-rank methods, in model performance, computation complexity and eigenvalues spectrum analysis in \Cref{fig:eigenspectrum} and \Cref{tab:lowrank}. 

\begin{table}
    \centering
    \caption{Model performance and computation complexity between low-rank methods and K-DCT.} \label{tab:lowrank}
    \begin{tabular}{lccc}
        \toprule
         Covariance model &	FID &	Memory (MB) & Time (ms)\\
         \midrule 
         \multicolumn{4}{l}{CIFAR10 (LS)}\\
         NPR-diag	& 32.35	& 272.39 &	417.66\\
         LowRank(r=5) &	28.81 &	320.46 &	419.68\\
         LowRank(r=10)	&27.57	&350.52	&420.52\\
         LowRank(r=25)	&26.79	&440.72&	425.36\\
         LowRank(r=50)	&26.14&	591.05&	428.15\\
         NPR-KDCT	&23.06	&298.59&	420.67\\
         \midrule
         \multicolumn{4}{l}{CelebA}\\
         NPR-diag &	28.37	&552.05&	1000.55\\
        LowRank(r=5)&	26.15&	768.11&	1017.01\\
        LowRank(r=25)&	24.18&	1224.38&	1026.02\\
        LowRank(r=50)&	23.35&	1848.71&	1039.35\\
        LowRank(r=100)&	23.16&	3050.05&	1061.58\\
        NPR-KDCT&	19.69&	676.81&	1146.25 \\
        \bottomrule
    \end{tabular}
\end{table}

\subsection{Ablation studies on the DCT basis} \label{sec:ablation}

As shown in \Cref{fig:kronecker-celeba}, the DCT basis fits CelebA worse than ImageNet, while the K-DCT structure still shows improvements both in quality and likelihood estimate for the generated samples on the CelebA dataset. 
We set out to test whether fixing the spatial eigenbasis to DCT is harmful by two ablation experiments: 
\begin{wraptable}{r}{8cm}
    \centering
    \caption{What does the DCT matrix bring? \label{tab:ablations}}
    \begin{tabular}{lccc}
        \toprule
         Model &	\multicolumn{3}{c}{FID} \\
         & $K=10$ & $K=25$ & $K=50$ \\
         \midrule 
         DCT & 17.51 & 10.88 & 7.57 \\
         FreeDiag & 17.49 & 10.80 & 7.52 \\
         Free & 16.97 & 10.60 & 7.47 \\
        \bottomrule
    \end{tabular}
    \vspace{-1.2cm}
\end{wraptable}
\begin{itemize}
    \item First (labelled ``Free'' in \Cref{tab:ablations}), we relaxed the spatial eigenbasis ($F \otimes F$ in Eq.~\ref{eq:our_param}) to two learnable matrices $A$ and $B$ acting on horizontal and vertical image dimensions respectively, resulting in a covariance model of the form $\mathcal{E}_\phi = \text{diag}(\epsilon_\phi) + C_\phi C_\phi^\top \otimes (A \otimes B)^\top \text{diag}(\lambda_\phi) (A \otimes B)$. 
    \item Second (labelled ``FreeDiag''), we only partially relaxed the eigenbasis by allowing for diagonal rescaling of the DCT matrix without otherwise affecting its main structure. This takes the form $\mathcal{E}_\phi = \text{diag}(\epsilon_\phi) + C_\phi C_\phi^\top \otimes (AFA \otimes BFB)^\top \text{diag}(\lambda_\phi) (AFA \otimes BFB)$, where $F$ is still the fixed DCT operator, while $A$ and $B$ are small diagonal matrices that modulate $F$ along the horizontal and vertical image dimensions and potentially capture some lack of translation invariance in the data.
\end{itemize} 

We trained both relaxations for 50k iterations; for ``Free'', we initialized both $A$ and $B$ to be the same as $F$, and made them trainable thereafter; for ``FreeDiag'', we initialized $A$ and $B$ to be identity matrices.

Whilst both relaxations bring some improvements on FID (in the direction one would expect: ``Free'' better than ``FreeDiag'' better than our original KDCT formulation), these improvements are relatively modest -- certainly much smaller than the improvement made by the original KDCT over a purely diagonal model.

\begin{figure}
    \centering
    \includegraphics[width=0.7\linewidth]{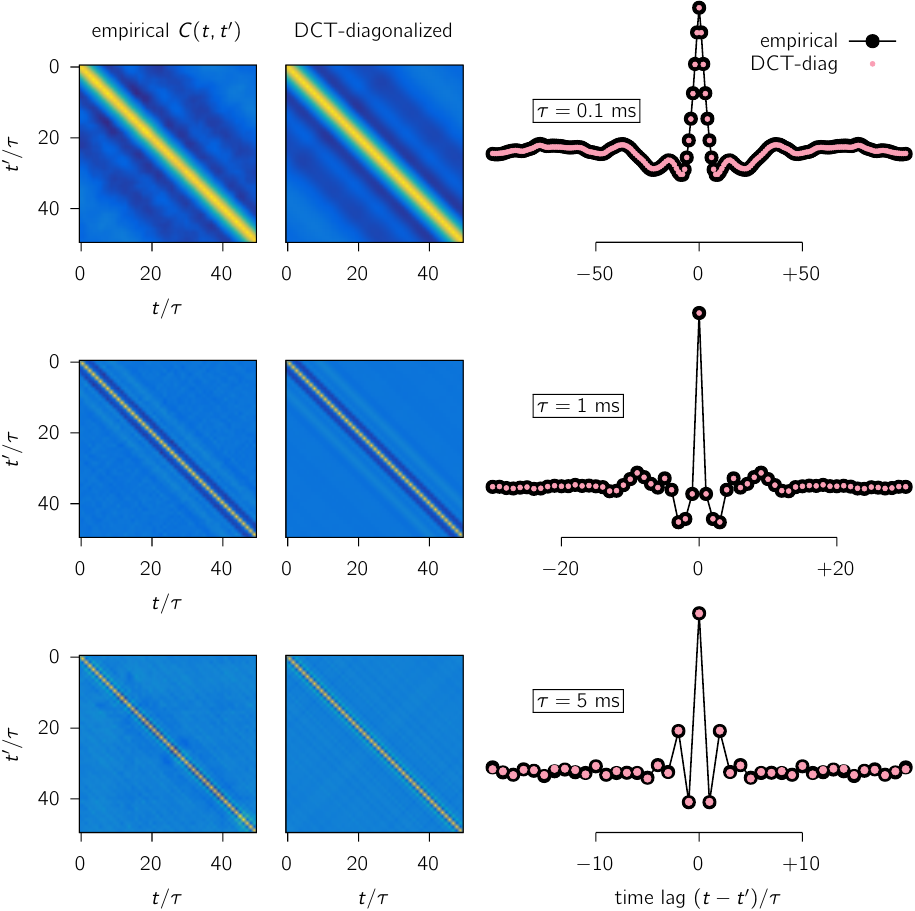}
    \caption{\textbf{Accuracy of DCT-based parameterization for audio (speech) data.}
    \textbf{Left:}~Empirical covariance matrix $\Sigma$ of speech (LibriSpeech dataset) at $\tau=0.1$~ms resolution (top), $1$~ms (middle) and $5$~ms (bottom). 
    \textbf{Center:}~Nearest DCT-diagonalized approximation, $\Sigma \approx \hat\Sigma \equiv F^\top D F $ where $F$ is the DCT matrix and $D$ is the diagonal matrix that minimizes $\| \Sigma - \hat\Sigma \|_\text{F}^2$.
    \textbf{Right:}~Marginal slices along the secondary diagonal, which collapse those covariance matrices into covariance as a function of time lag $(t-t^\prime)/\tau$ assuming stationarity. 
    The DCT parameterization provides a very good fit to the empirical covariance at all resolutions.
    (Due to limited number of samples in the dataset, covariances on slower timescales ($\tau > 10$~ms) could not be estimated as accurately.)}
     \label{fig:suppl:audio}
\end{figure}
\end{document}

%% file: math_commands.tex
\usepackage{amsmath,amsfonts,bm}

\def\eqref#1{equation~\ref{#1}}

\def\1{\bm{1}}

\DeclareMathAlphabet{\mathsfit}{\encodingdefault}{\sfdefault}{m}{sl}
\SetMathAlphabet{\mathsfit}{bold}{\encodingdefault}{\sfdefault}{bx}{n}



%% file: tables/fid-nll.tex
\begin{tabular}{lrrrrrrrrrrrr}
    \toprule
     \textbf{FID} & \multicolumn{3}{c}{CIFAR10~(LS)}                   & \multicolumn{3}{c}{CIFAR10~(CS)} & \multicolumn{3}{c}{CelebA $64\times64$}                   & \multicolumn{3}{c}{ImageNet $64\times64$}  \\
    \cmidrule(r){2-4} \cmidrule(r){5-7} \cmidrule(r){8-10} \cmidrule(r){11-13}
    \# Timesteps $K$     & 10 & 25 & 50 & 10 & 25 & 50 & 10 & 25 & 50 & 25 & 50 & 100 
    \\
    \midrule
    Heuristic $\tilde\beta_t$ & 44.45 & 21.83 & 15.21 & 34.76 & 16.18 & 11.11 & 36.69 & 24.46 & 18.96 & 29.21 & 21.71 & 19.12\\
    Heuristic $\beta_t$ & 233.41 & 125.05 & 66.28 & 205.31 & 84.71 & 37.35 & 294.79 & 115.69 & 53.39 & 170.28 & 83.86 & 45.04 \\
    SN-diagonal & 24.06 & \textcolor{Bittersweet}{\textbf{6.91}} & \textcolor{Bittersweet}{\textbf{4.63}} & 16.33 & \textcolor{NavyBlue}{\textbf{6.05}} & \textcolor{NavyBlue}{\textbf{4.17}} & 20.60 & \textcolor{NavyBlue}{\textbf{12.00}} & \textcolor{NavyBlue}{\textbf{7.88}} & 27.58 & 20.74 & 18.04 \\
    NPR-diagonal & 32.35 & 10.55 & 6.18 & 19.94 & 7.99 & 5.31 &28.37 &15.74 &10.89 & 28.27 & 20.89 & 18.06  \\
    
    \rowcolor{gray!10}
    NPR-K-DCT (ours) & \textcolor{NavyBlue}{\textbf{23.06}} & 9.12 & 5.92 & 17.26 & 7.79 & 5.58 & \textcolor{NavyBlue}{\textbf{19.69}} & 13.72 & 9.80 & \textcolor{NavyBlue}{\textbf{23.98}}  &\textcolor{NavyBlue}{\textbf{18.90}} & \textcolor{NavyBlue}{\textbf{17.44}} \\
    
    OCM-diagonal & 24.94 & 9.19 & 5.95 & \textcolor{NavyBlue}{\textbf{14.32}} & \textcolor{Bittersweet}{\textbf{5.54}} & \textcolor{Bittersweet}{\textbf{4.10}} & 21.55 & 12.71 & 9.24 & 28.02 &20.81 & 17.98  \\
    
    \rowcolor{gray!10}
    OCM-K-DCT (ours) & \textcolor{Bittersweet}{\textbf{22.30}} & \textcolor{NavyBlue}{\textbf{9.10}} & \textcolor{NavyBlue}{\textbf{5.92}} &  \textcolor{Bittersweet}{\textbf{\textbf{12.96}}} &6.28 &5.05 & \textcolor{Bittersweet}{\textbf{17.51}} & \textcolor{Bittersweet}{\textbf{10.88}} & \textcolor{Bittersweet}{\textbf{7.57}} & \textcolor{Bittersweet}{\textbf{23.88}} & \textcolor{Bittersweet}{\textbf{18.78}} & \textcolor{Bittersweet}{\textbf{17.39}} \\
    
    \midrule
    
    \textbf{NLL} $\approx$ -ELBO & \multicolumn{3}{c}{CIFAR10~(LS)}                   & \multicolumn{3}{c}{CIFAR10~(CS)} & \multicolumn{3}{c}{CelebA $64\times64$}                   & \multicolumn{3}{c}{ImageNet $64\times64$} \\
    \cmidrule(r){2-4} \cmidrule(r){5-7} \cmidrule(r){8-10} \cmidrule(r){11-13}
    \rowcolor{white}
    \# Timesteps $K$    & 10 & 25 & 50 & 10 & 25 & 50  & 10 & 25 & 50 & 25 & 50 & 100 
    \\
    \midrule
    Heuristic $\tilde\beta_t$ & 74.95 &24.98 &12.01 & 75.96 &24.94 &11.96& 33.42 &13.09 &7.14 &105.87 &46.25 &22.02\\ 
    Heuristic $\beta_t$ & 6.99 &6.11 &5.44 & 6.51 &5.55 &4.92 & 6.67 &5.72 &4.98 & 5.81 &5.20 &4.70\\
    SN-diagonal & 30.79 & 11.83 & 7.13 &  90.85 & 19.81 & 9.72 & 18.09 & 8.05 & 5.29 & 4.56 & 4.18 & 3.95 \\
    NPR-diagonal & 5.40 &4.64 &4.25 & 5.03 &4.33 &3.99  &4.46 &3.78 &3.40 & 4.66 &4.22 &3.96 \\
    
    \rowcolor{gray!10}
    NPR-K-DCT (ours) & \textcolor{Bittersweet}{\textbf{4.50}} & \textcolor{Bittersweet}{\textbf{4.10}} & \textcolor{Bittersweet}{\textbf{3.90}} & \textcolor{Bittersweet}{\textbf{4.31}} &\textcolor{Bittersweet}{\textbf{3.98}}  & \textcolor{Bittersweet}{\textbf{3.87}} & \textcolor{NavyBlue}{\textbf{3.45}} & \textcolor{NavyBlue}{\textbf{3.17}} & \textcolor{NavyBlue}{\textbf{2.99}} & \textcolor{Bittersweet}{\textbf{4.29}}& \textcolor{Bittersweet}{\textbf{4.07}}& \textcolor{Bittersweet}{\textbf{3.91}} \\
    OCM-diagonal & 5.32 &4.63 &4.25 & 4.99 &4.34 &3.99 & 4.69 &3.86 &3.43 & 4.45 &4.15 &3.93\\
    
    \rowcolor{gray!10}
    OCM-K-DCT (ours) & \textcolor{NavyBlue}{\textbf{4.61}} & \textcolor{NavyBlue}{\textbf{4.17}} & \textcolor{NavyBlue}{\textbf{4.02}}& \textcolor{NavyBlue}{\textbf{4.82}} &\textcolor{NavyBlue}{\textbf{4.24}} & \textcolor{NavyBlue}{\textbf{3.94}} & \textcolor{Bittersweet}{\textbf{3.44}} & \textcolor{Bittersweet}{\textbf{3.16}} & \textcolor{Bittersweet}{\textbf{2.99}} & \textcolor{NavyBlue}{\textbf{4.41}} & 
    \textcolor{NavyBlue}{\textbf{4.14}}& 
    \textcolor{NavyBlue}{\textbf{3.93}}\\
    \bottomrule
\end{tabular}

%% file: tables/std-mean.tex
\begin{tabular}{lrrrrrr}
    \toprule
     \textbf{FID} & \multicolumn{3}{c}{CIFAR10~(LS)}                   & \multicolumn{3}{c}{CIFAR10~(CS)} \\
    \arrayrulecolor{black!30} \cmidrule(r){2-4} \cmidrule(r){5-7}
    \# Timesteps $K$     & 10 & 25 & 50 & 10 & 25 & 50 
    \\
    \arrayrulecolor{black}\midrule
    MEAN (K-DCT-NPR) & 22.921 &	9.142 &	5.880 & 17.545 & 7.661 &	5.568
    \\
    STD (K-DCT-NPR) & 0.154 &	0.019 &	0.040 & 0.032 &	0.108 &	0.011
    \\
    \arrayrulecolor{black!30}\cmidrule(r){2-7}
    MEAN (K-DCT-OCM) & 21.980 &	9.082 &	5.914 & 12.865 & 6.298 &	5.092
    \\
    STD (K-DCT-OCM) & 0.291 &	0.022 &	0.022 & 0.132 &	0.026 & 0.042
    \\
    \arrayrulecolor{black}\midrule
    & \multicolumn{3}{c}{CelebA $64\times64$}                   & \multicolumn{3}{c}{ImageNet $64\times64$}
    \\
    \arrayrulecolor{black!30} \cmidrule(r){2-4} \cmidrule(r){5-7}
    \# Timesteps $K$     & 10 & 25 & 50 & 25 & 50 & 100
    \\
    \arrayrulecolor{black}\midrule
    MEAN (K-DCT-NPR) & 19.773 &	13.727 & 9.842 & 23.914 &	18.796 & 17.315
    \\
    STD (K-DCT-NPR) & 0.090 &	0.080 &	0.038 & 0.093 &	0.131 &	0.135
    \\
    \arrayrulecolor{black!30}\cmidrule(r){2-7}
    MEAN (K-DCT-OCM) & 17.457 &	10.873 & 7.447 & 23.812 & 18.833 &	17.262
    \\
    STD (K-DCT-OCM) & 0.063 &	0.004 &	0.137 & 0.109 &	0.132 &	0.128
    \\
    \arrayrulecolor{black}\bottomrule
    \\
    \textbf{NLL} & \multicolumn{3}{c}{CIFAR10~(LS)}                   & \multicolumn{3}{c}{CIFAR10~(CS)} \\
    \arrayrulecolor{black!30} \cmidrule(r){2-4} \cmidrule(r){5-7}
    \# Timesteps $K$     & 10 & 25 & 50 & 10 & 25 & 50 
    \\
    \arrayrulecolor{black}\midrule
    MEAN (K-DCT-NPR) & 4.4983 &	4.1076 &	3.9051 & 4.3017 & 	3.9853 & 3.8575
    \\
    STD (K-DCT-NPR) & 0.0028 &	0.0101 &	0.0043 & 0.0032 &	0.0027 &	0.0093
    \\
    \arrayrulecolor{black!30}\cmidrule(r){2-7}
    MEAN (K-DCT-OCM) & 4.6079 &	4.1757 &	4.0222 & 4.8201 &	4.2421 & 3.9463
    \\
    STD (K-DCT-OCM) & 0.0030 &	0.0045 &	0.0026 & 0.0052 &	0.0053 &	0.0054
    \\
    \arrayrulecolor{black}\midrule
    & \multicolumn{3}{c}{CelebA $64\times64$}                   & \multicolumn{3}{c}{ImageNet $64\times64$}
    \\
    \arrayrulecolor{black!30} \cmidrule(r){2-4} \cmidrule(r){5-7}
    \# Timesteps $K$     & 10 & 25 & 50 & 25 & 50 & 100
    \\
    \arrayrulecolor{black}\midrule
    MEAN (K-DCT-NPR) & 3.4480 &	3.1714 & 2.9928 & 4.2986 &	4.0770 &	3.9044
    \\
    STD (K-DCT-NPR) & 0.000649 &	0.000520 &	0.000147 & 0.000406 &	0.000078 &	0.000367
    \\
    \arrayrulecolor{black!30}\cmidrule(r){2-7}
    MEAN (K-DCT-OCM) & 3.4383 &	3.1663 &	2.9907 & 4.4180 &	4.1446 &	3.9318
    \\
    STD (K-DCT-OCM) & 0.000137 &	0.000073 &	0.000086 & 0.000093 & 0.000015 & 	0.000051
    \\
    \arrayrulecolor{black}\bottomrule
\end{tabular}